\documentclass[11pt]{article}
\usepackage[preprint]{acl}
\usepackage{times}
\usepackage{latexsym}
\usepackage{booktabs}
\usepackage[T1]{fontenc}
\usepackage[utf8]{inputenc}
\usepackage{microtype}
\usepackage{inconsolata}
\usepackage{graphicx}
\usepackage{amssymb}
\usepackage{amsmath}
\usepackage{amsthm}
\newtheorem{thm}{Theorem}
\usepackage{multirow}
\usepackage{adjustbox}
\title{TeRA: Vector-based Random Tensor Network for High-Rank Adaptation of Large Language Models}

\author{
 \textbf{Yuxuan Gu\thanks{Equal Contribution.}\textsuperscript{1}},
 \textbf{Wuyang Zhou\footnotemark[1]\thanks{Project Lead.}\textsuperscript{1}},
 \textbf{Giorgos Iacovides\textsuperscript{1}},
 \textbf{Danilo Mandic\textsuperscript{1}}
\\
\\
 \textsuperscript{1}Department of Electrical and Electronic Engineering, Imperial College London\\ \{yuxuan.gu21, wuyang.zhou19, giorgos.iacovides20, d.mandic\}@imperial.ac.uk
}

\begin{document}
\maketitle
\begin{abstract}
Parameter-Efficient Fine-Tuning (PEFT) methods, such as Low-Rank Adaptation (LoRA), have significantly reduced the number of trainable parameters needed in fine-tuning large language models (LLMs). The developments of LoRA-style adapters have considered two main directions: (1) enhancing model expressivity with high-rank adapters, and (2) aiming for further parameter reduction, as exemplified by vector-based methods. However, these approaches come with a trade-off, as achieving the expressivity of high-rank weight updates typically comes at the cost of sacrificing the extreme parameter efficiency offered by vector-based techniques. To address this issue, we propose a vector-based random \underline{\textbf{Te}}nsor network for high-\underline{\textbf{R}}ank \underline{\textbf{A}}daptation (TeRA), a novel PEFT method that achieves high-rank weight updates while retaining the parameter efficiency of vector-based PEFT adapters. This is achieved by parametrizing the tensorized weight update matrix as a Tucker-like tensor network (TN), whereby large randomly initialized factors are frozen and shared across layers, while only small layer-specific scaling vectors, corresponding to diagonal entries of factor matrices, are trained.  Comprehensive experiments demonstrate that TeRA matches or even outperforms existing high-rank adapters, while requiring as few trainable parameters as vector-based methods. Theoretical analysis and ablation studies validate the effectiveness of the proposed TeRA method. The code is available at \url{https://github.com/guyuxuan9/TeRA}.
\end{abstract}

\section{Introduction}

Foundation models, such as the Llama \cite{touvron2023llama,grattafiori2024llama} and GPT \cite{brown2020language, achiam2023gpt} series, have revolutionized the field of natural language processing (NLP) by demonstrating strong generalization abilities across a diverse range of tasks. Although these models have been pre-trained on a large-scale corpus of textual data, supervised fine-tuning (SFT) is often necessary to improve their performance on specific downstream tasks. However, the large number of trainable parameters in full-parameter SFT can become computationally prohibitive in resource-constrained environments. 

\begin{figure}[t]
    \centering
    \includegraphics[width=0.48\textwidth]{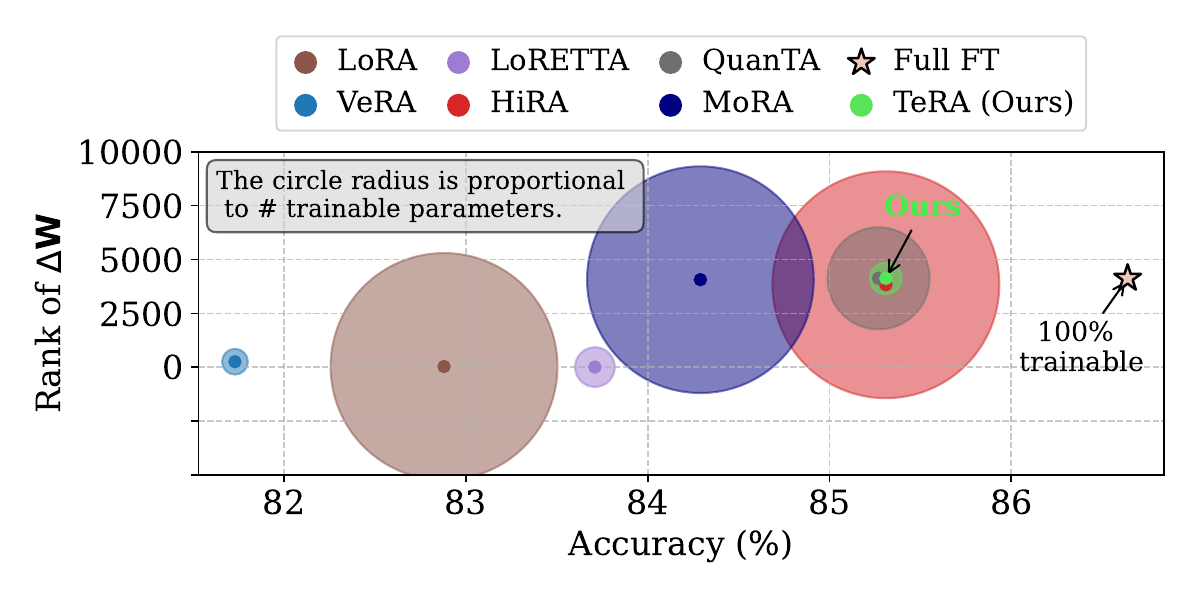}
    \caption{TeRA exhibits superior performance, high-rank and parameter efficiency trade-off. On commonsense reasoning task with Llama-3-8B, TeRA achieves high-rank updates and great performance, while keeping as few trainable parameters as vector-based methods (VeRA). Full Fine-tuning (FFT) requires many parameters and its parameter count is omitted (See Table \ref{tab:common_sense}).}
    \label{fig:comparison}
\end{figure}

Parameter-Efficient Fine-Tuning (PEFT) methods, particularly Low-Rank Adaptation (LoRA) \cite{hu2022lora}, have substantially reduced the number of trainable parameters in adapting large language models (LLMs) to downstream tasks. These methods operate by freezing the original pre-trained weights and training only the weight updates, $\Delta \mathbf{W} \in \mathbb{R}^{J_1 \times J_2}$, which are used to modify the original weights. The core assumption of LoRA is that these weight updates can be effectively approximated by a low-rank decomposition, $\Delta \mathbf{W}=\mathbf{A}\mathbf{B}$, where $\mathbf{A} \in \mathbb{R}^{J_1 \times r}$ and $\mathbf{B} \in \mathbb{R}^{r \times J_2}$, where $r$ controlls the rank upper bound.

\begin{figure*}[!t]
    \centering
    \includegraphics[width=\textwidth]{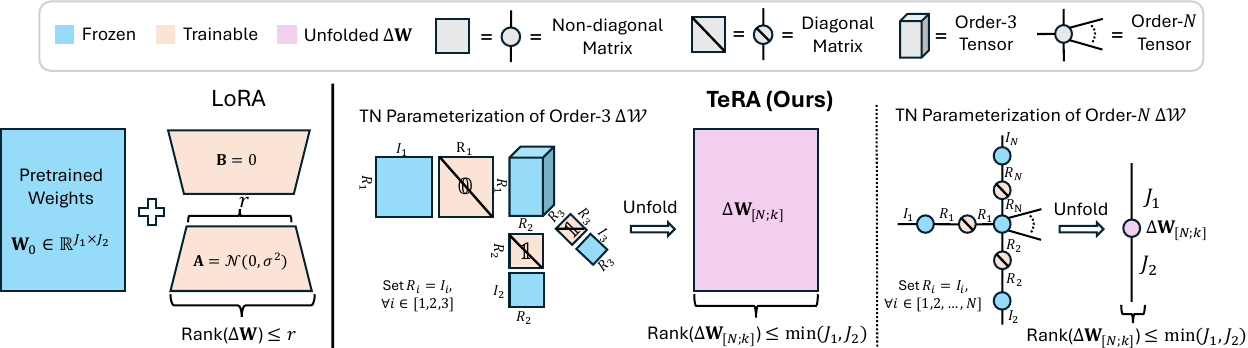}
    \caption{A comparison between the matrix-based LoRA \cite{hu2022lora} and our proposed tensor-based TeRA methods. LoRA represents the weight update matrix using two smaller matrices, while TeRA employs a Tucker-like \cite{tucker1964extension} multi-linear tensor network (TN) to parametrize the tensorized $\Delta \mathcal{W}$. This design allows TeRA to achieve high-rank updates with much fewer trainable parameters than LoRA.}
    \label{fig:architecture}
\end{figure*}

However, the low rank assumption of $\Delta \mathbf{W}$ can limit its expressivity when adapting to more complex downstream tasks \cite{jiang2024morahighrankupdatingparameterefficient, huang2025hira}. To address this, \citet{huang2025hira} proposed Hadamard High-Rank Adaptation (HiRA), which enhances update expressivity by introducing a Hadamard product between the learned weight update matrix, $\mathbf{A} \mathbf{B}$, and the frozen pre-trained weight matrix, $\mathbf{W}_0$. Specifically, HiRA defines the high-rank weight update as $\Delta \mathbf{W} = (\mathbf{A} \mathbf{B}) \odot \mathbf{W}_0$, where $\odot$ denotes the Hadamard product, and $\mathbf{W}_0$ is the frozen pre-trained weight matrix. Similar to LoRA, HiRA still requires training all parameters in $\mathbf{A}$ and $\mathbf{B}$.

Methods like VeRA \cite{kopiczko2024vera} reduce the trainable parameter count further by freezing $\mathbf{A}$ and $\mathbf{B}$, while training only two scaling vectors, $\mathbf{b}$ and $\mathbf{d}$, but they remain constrained by the low-rank assumption. Specifically, the VeRA weight update is parametrized as $\Delta \mathbf{W} = \mathbf{\Lambda}_b \mathbf{B} \mathbf{\Lambda}_d \mathbf{A}$, where $\mathbf{b} \in \mathbb{R}^{J_1}$ and $\mathbf{d} \in \mathbb{R}^{r}$ are the two trainable diagonal entries in $\mathbf{\Lambda}_b \in \mathbb{R}^{J_1 \times J_1}$ and $\mathbf{\Lambda}_d \in \mathbb{R}^{r \times r}$, respectively. Thus, VeRA requires only a fraction of the trainable parameters of LoRA.

It is possible to obtain high-rank weight updates with low-rank adapters. However, this leads to an explosion in the number of trainable parameters. Meanwhile, the low-rank assumption in the weight update matrices has been shown to restrict their performance in more complex tasks, such as arithmetic and reasoning \cite{huang2025hira}. It is therefore natural to ask:

\begin{itemize}
    \item \textit{Is it possible to achieve high- (or full-) rank weight updates and  their desired performance, while maintaining a similar number of trainable parameters to vector-based PEFT adapters such as VeRA?}
\end{itemize}

To resolve this trade-off, we propose a vector-based random \underline{\textbf{Te}}nsor network for high-\underline{\textbf{R}}ank \underline{\textbf{A}}daptation (TeRA), a novel PEFT method that achieves high-rank weight updates using very few trainable parameters. The core idea is to tensorize the weight update matrix, $\Delta \mathbf{W}$, into a higher-order tensor, and then parametrize it using a Tucker-like \cite{tucker1964extension} tensor network \cite{7038247}, as shown in Figure \ref{fig:architecture}. Within this tensor network, we freeze large randomly initialized factors, share them across all layers, and train only the small layer-specific scaling vectors that parametrize diagonal factor matrices. This design effectively decouples the rank of the update matrix from the number of trainable parameters, enabling high-rank adaptation with extremely low parameter counts similar to vector-based methods.

\begin{figure}[t]
    \centering
    \includegraphics[width=0.48\textwidth]{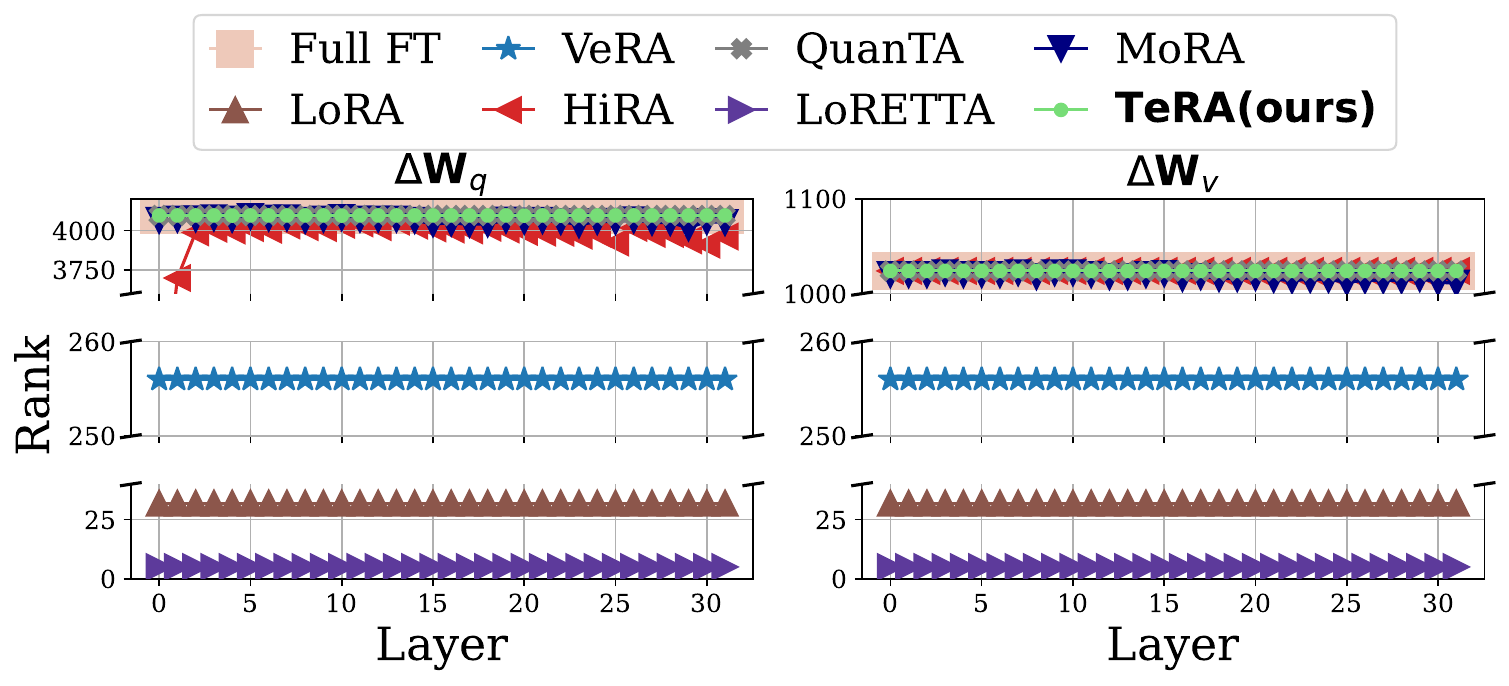}
    \caption{Rank analysis of $\Delta \mathbf{W}_q$ (max allowed rank of $4096$) and $\Delta \mathbf{W}_v$ (max allowed rank of $1024$) across Llama-3-8B layers. TeRA consistently maintains a high (near-full) rank. In contrast, methods like LoRA and VeRA have lower-rank weight updates, limiting their expressivity. See Figure \ref{fig:rank_highrank_methods_appen} in Appendix \ref{sec:rank_com} for zoomed-in comparison of high-rank adapters.}
    \label{fig:rank_different_methods}
\end{figure}

Extensive experiments demonstrate that TeRA establishes a superior trade-off between model performance, high rank, and parameter efficiency. As shown in Figure \ref{fig:comparison}, TeRA matches the accuracy of HiRA using orders of magnitude fewer parameters. Compared to parameter count-matched methods like VeRA and LoRETTA \cite{yang2024loretta}, TeRA offers a significant accuracy improvement. This stems from the more expressive high-rank weight updates in TeRA across all model layers (See Figure \ref{fig:rank_different_methods}), a property that low-rank methods inherently lack. As shown, TeRA maintains near full-rank updates for $\Delta \mathbf{W}_q$ and $\Delta \mathbf{W}_v$ across all layers, enabling more expressive adaptations.

In summary, our contributions are as follows:
\begin{itemize}
    \item We propose TeRA, a new PEFT method that uses a Tucker-like tensor network to parametrize the tensorized high-rank weight updates, which can be merged with the original weights and incur zero inference overhead.
    \item A theoretical analysis is provided demonstrating that TeRA can achieve high-rank weight updates with fewer parameters than existing methods. Our analysis formalizes the trade-off between the performance and the trainable parameter count of TeRA.
    \item Extensive experiments compare TeRA with baseline methods, demonstrating that TeRA exhibits superior performance while requiring a similar number of trainable parameters to vector-based PEFT adapters.
\end{itemize}

\section{Related work}

\paragraph{Prompt-based Methods.} One category of PEFT methods comprises prompt-based methods, such as Prompt Tuning \cite{lester-etal-2021-power} and P-Tuning \cite{liu-etal-2022-p}, which introduce additional trainable virtual tokens into the input of LLMs and optimize only these tokens. These methods are sensitive to initialization schemes and require additional computational costs during inference.

\paragraph{Low-rank Adaptation (LoRA).} Introduced by \citet{hu2022lora}, LoRA employs two matrices, $\mathbf{A} \in \mathbb{R}^{J_1 \times r} $ and $\mathbf{B} \in \mathbb{R}^{r \times J_2}$, to parametrize the weight update matrix, $\Delta \mathbf{W} = \mathbf{A} \times \mathbf{B} \in \mathbb{R}^{J_1 \times J_2}$, as a low-rank decomposition, thereby significantly reducing the number of trainable parameters. Since $\Delta \mathbf{W}$ has the same dimensionality as the pre-trained weight matrix, it can be merged into the pre-trained weight during inference, eliminating any additional inference overhead. Based on LoRA, VeRA \cite{kopiczko2024vera} proposes to randomly initialize and freeze the $\mathbf{A}$ and $\mathbf{B}$ matrices, share them across layers, and train only two scaling vectors, $\mathbf{b}$ and $\mathbf{d}$, thus significantly reducing the trainable parameters needed for low-rank  weight updates. Other variants of LoRA impose different algebraic structures within $\Delta \mathbf{W}$ to achieve parameter reduction or high-rank updates \cite{zhang2023adaptive, dettmers2023qlora, liu2024dora, borse2024foura}.

Recently, tensor-based methods, which operate on tensorized neural network weights \cite{gu2025tensorllm, zhou2026kromhc}, have also been shown to be effective in fine-tuning LLMs \cite{yang2024loretta,bershatsky2024lotr}. More specifically, they focus on reducing the number of trainable parameters compared to LoRA by assuming higher-order low-rank update structures \cite{ttd}, which may still fail to capture high-rank updates for complex tasks \cite{huang2025hira}. \par

\paragraph{High-rank Adaptation.} To overcome the limited expressivity of low-rank adaptation, high-rank variants of LoRA have been proposed. QuanTA \cite{chen2024quanta} uses a quantum-inspired method to efficiently achieve high-rank weight updates. MoRA \cite{jiang2024morahighrankupdatingparameterefficient} employs a square matrix to achieve high-rank updates, while HiRA \cite{huang2025hira} uses the Hadamard product to learn a high-rank weight update matrix. Different from these methods, TeRA effectively achieves high-rank weight updates and uses much fewer trainable parameters, similar amount to that in vector-based PEFT adapters such as VeRA.

\section{Tensors and Multi-linear Algebra Preliminaries}
The mathematical notations used in this paper are listed in Table \ref{tab:math_notation}. This is consistent with the notation used in \citet{7038247}.

\begin{table}[htbp]
\footnotesize
  \centering
  \begin{tabular}{cc}
    \toprule
      Symbol   & Meaning \\
    \midrule
    $a$, $\mathbf{a}$, $\mathbf{A}$, $\mathcal{A}$    & Scalar, vector, matrix, tensor  \\
    $(\cdot)^\top$ & Matrix transpose \\
    $(\cdot)^{\dagger}$ & Matrix pseudoinverse \\
    $\|\cdot\|_F$ & Frobenius norm \\
    $\mathcal{A}(i_1, \dots, i_N)$ & The $(i_1, \dots, i_N)$-th element of $\mathcal{A}$ \\
    $\mathcal{A} \times_n \mathbf{B}$ & Mode-$n$ product \\
    diag$\left(\mathbf{a}\right)$ & A diagonal matrix whose diagonal is $\mathbf{a}$ \\
    \bottomrule
    
  \end{tabular}
  \caption{Mathematical notations}
  \label{tab:math_notation}
\end{table}

A tensor is a multi-dimensional array and a higher-order generalization of vectors and matrices, whereby a vector, $\mathbf{a} \in \mathbb{R}^{I_1}$ is an order-$1$ tensor, while a matrix, $\mathbf{A} \in \mathbb{R}^{I_1 \times I_2}$, is an order-$2$ tensor. An order-$N$ tensor is denoted by $\mathcal{A} \in \mathbb{R}^{I_1\times I_2 \times \cdots \times I_N}$.

\paragraph{Tensorization and Matricization.}
Tensorization (folding) is the process to fold a lower-dimensional tensor into a higher-dimensional one. A matrix $\mathbf{A} \in \mathbb{R}^{J_1 \times J_2}$ can be folded into an order-$N$ tensor $\mathcal{A} \in \mathbb{R}^{I_1 \times \cdots \times I_N}$, provided that $\prod_{i=1}^{k} I_i = J_1$ and $\prod_{i=k+1}^{N} I_i = J_2$ for some $k \in [1,N]$. Its inverse operation is termed matricization (unfolding). Unfolding operation converts an order-$N$ tensor, $\mathcal{A} \in \mathbb{R}^{I_1\times \cdots \times I_N}$, into a matrix, $\mathbf{A}_{[N;k]}\in \mathbb{R}^{\prod_{i = 1}^k I_{i} \times \prod_{i = k+1}^N I_{i}}$, whose elements are given by the following
\begin{equation}
\begin{aligned}
&\mathbf{A}_{[N;k]}(\overline{i_{1} \cdots i_{k}}, \overline{i_{k+1} \cdots i_{N}}) = \mathcal{A}(i_{1}, i_{2}, \ldots, i_{N}), 
\end{aligned}
\end{equation}
The corresponding tensorization operation is denoted by $\text{Fold}_{[N;k]}(\mathbf{A}_{[N;k]}) = \mathcal{A}$. 

\paragraph{Mode-$n$ product.}
Mode-$n$ product between a tensor $\mathcal{A} \in \mathbb{R}^{I_1 \times I_2 \times \cdots \times I_N}$ and a matrix $\mathbf{B} \in \mathbb{R}^{I_n \times J_n}$ yields a tensor $\mathcal{C} \in \mathbb{R}^{I_1 \times \cdots \times I_{n-1} \times J_n \times I_{n+1} \times \cdots \times I_N}$. This operation is denoted as 
\begin{equation}
    \mathcal{C} = \mathcal{A} \times_n \mathbf{B},
\end{equation}
and its element-wise definition form is
\begin{equation}
\begin{aligned}
    &\mathcal{C}(i_1,\cdots, i_{n-1}, j_n, i_{n+1}, \cdots,i_N)= \\
    & \sum_{i_n=1}^{I_N} \mathcal{A}(i_1,\cdots,i_{n-1}, i_n,i_{n+1},\cdots,i_N) \mathbf{B}(i_n, j_n).
\end{aligned}
\end{equation}

\paragraph{Tucker Decomposition.}
Tucker decomposition \cite{tucker1964extension} is a generalization of Singular Value Decomposition (SVD) to higher-order tensors \cite{de2000multilinear} and a cornerstone of multi-linear tensor network \cite{7038247, zhou2024tensor}. Given an order-$N$ tensor, $\mathcal{X} \in \mathbb{R}^{I_1 \times I_2 \times \cdots \times I_N}$, Tucker decomposition expresses it using a smaller order-$N$ core tensor $\mathcal{G} \in \mathbb{R}^{R_1 \times R_2 \times \cdots \times R_N}$, where $R_i \ll I_i$, and $N$ factor matrices $\{\mathbf{B}^{(i)}  \in \mathbb{R}^{R_i \times I_i}\}_{i=1}^N$, in the form
\begin{equation}
    \mathcal{X} = \mathcal{G} \times_1 \mathbf{B}^{(1)} \times_2 \mathbf{B}^{(2)} \times_3 \cdots \times_N \mathbf{B}^{(N)}.
\end{equation}
Identifying the optimal set of Tucker ranks, $[R_1, \dots, R_N]$, efficiently is an active area of research \cite{zhou2025understanding}, with numerous recent studies focusing on advanced methods for tensor rank search \cite{iacovides2024towards, tnale_cite, iacovides2025domain}. \\

\section{Methodology}
TeRA parametrizes the tensorized weight update matrix $\Delta \mathbf{W}$ using a Tucker-like tensor network, as shown in Figure \ref{fig:architecture}. The method involves two steps: (1) Tensorize the weight update matrix into a higher-order tensor $\Delta \mathcal{W} \in \mathbb{R}^{I_1 \times \cdots \times I_N}$; (2) Parametrize this tensor using a Tucker-like tensor network with majority of large factor matrices frozen, and only small diagonal matrices trainable. After training, $\Delta \mathcal{W}$ is unfolded to obtain $\Delta \mathbf{W}_{[N;k]}$.

Specifically, TeRA tensorizes a weight update matrix into an order-$N$ tensor $\Delta\mathcal{W} \in \mathbb{R}^{I_1 \times I_2 \times \cdots \times I_N} =\operatorname{Fold}_{[N;k]}(\Delta \mathbf{W}_{[N;k]} \in \mathbb{R}^{J_1 \times J_2})$, where $\prod_{i = 1}^k I_{i} = J_1, \ \prod_{i = k+1}^N I_{i} = J_2$, $I_i \geq 2 \ \forall \ i=1,\ldots,N$, and $1\leq k < N$. For example, the attention weight matrices in Llama-2-7B have size $4096\times 4096$, which can be tensorized into shapes of $64 \times 64 \times 64 \times 64$, $16 \times 16 \times \cdots \times 16$, $4 \times 4 \times \cdots \times 4$, etc.

\paragraph{TeRA Formulation.} TeRA parametrizes the weight update tensor, $\Delta \mathcal{W}$, as the mode-$n$ product of a frozen core tensor $\mathcal{G}\in \mathbb{R}^{R_1 \times R_2 \times \cdots \times R_N}$, a set of $N$ frozen non-diagonal factor matrices $\{\mathbf{A}^{(i)}\in \mathbb{R}^{R_i \times I_i}\}_{i=1}^{N}$, and a set of $N$ trainable vectors $\{\mathbf{d}^{(i)} \in \mathbb{R}^{R_i}\}_{i=1}^N$, which are diagonal entries of $\{\operatorname{diag}(\mathbf{d}^{(i)})\in \mathbb{R}^{R_i \times R_i}\}_{i=1}^N$. The resulting parametrization is given by
\begin{equation}
\begin{aligned}
\label{eq:our_method}
\Delta \mathcal{W} = \mathcal{G} &\times_1 \operatorname{diag}(\mathbf{d}^{(1)}) \times_2 \operatorname{diag}(\mathbf{d}^{(2)}) \times_3 \\
& \cdots \times_N \operatorname{diag}(\mathbf{d}^{(N)}) \times_1 \mathbf{A}^{(1)}\\
& \times_2 \mathbf{A}^{(2)} \times_3 \cdots \times_N \mathbf{A}^{(N)}.
\end{aligned} 
\end{equation}

During fine-tuning, the core tensor $\mathcal{G}$ and the factor matrices $\{\mathbf{A}^{(i)}\}_{i=1}^N$ are randomly initialized and kept frozen. These matrices are shared between all the adapted layers of the LLM. The only trainable components are the diagonal entries of the matrices $\{\operatorname{diag}(\mathbf{d}^{(i)})\}_{i=1}^N$. All $\operatorname{diag}(\mathbf{d}^{(i)})$ are initialized as identity matrices, except for one $\operatorname{diag}(\mathbf{d}^{(i)})$, which is initialized as a zero matrix to ensure $\Delta \mathbf{W}_{[N;k]}$ is zero at initialization. This reduces the number of trainable parameters to just $\sum_{i=1}^{N} R_i$ per TeRA adapter, where the ranks $[R_1, \dots, R_N]$ are hyperparameters. 

Notably, TeRA introduces zero computational overhead during inference. After fine-tuning, the TeRA weight update $\Delta \mathbf{W}_{[N;k]}$ is unfolded from $\Delta \mathcal{W}$ and added to the pre-trained weights, $\mathbf{W}_0$, to yield the final weights
\begin{equation}
\mathbf{W}_{\text{final}} = \mathbf{W}_0 + \Delta \mathbf{W}_{[N;k]}.
\end{equation}

\begin{thm}
\label{thm:full-rank}
Let $\Delta \mathbf{W} \in \mathbb{R}^{J_1 \times J_2}$ be the weight update matrix, and $\Delta \mathcal{W} \in \mathbb{R}^{I_1 \times I_2 \times \cdots \times I_N} = \operatorname{Fold}_{[N;k]}(\Delta \mathbf{W}_{[N;k]} )$ be its folded weight update tensor, parametrized by TeRA as in Eq. (\ref{eq:our_method}). The following inequality holds
\begin{equation}
\operatorname{rank}\left(\Delta \mathbf{W}_{[N;k]}\right) \leq \operatorname{min}\left(\prod_{i=1}^k R_i, \prod_{i=k+1}^N R_i\right).
\end{equation} 
This allows for a full-rank update matrix under any tensorization (folding) schemes if $R_i = I_i \ \forall i = 1,\ldots,N$, i.e., $\operatorname{rank}\left(\Delta \mathbf{W}_{[N;k]}\right)  \leq \operatorname{min}\left(J_1, J_2\right)$ .
\end{thm}

This shows that TeRA not only can enable high-rank adaptation, but also requires a very small number of trainable parameters. For example, to allow for a full-rank weight update matrix of size $J_1 \times J_2$ ($J_1 \geq J_2$), we need $J_1\cdot J_2 + J_2\cdot J_2$ trainable parameters in LoRA and at \textit{least} $J_1+J_2$ trainable parameters in both VeRA and HiRA. However, TeRA only requires $\sum_{i=1}^N I_i $ trainable parameters, whereby $\prod_{i=1}^k I_i = J_1, \ \prod_{i=k+1}^N I_i = J_2$, $I_i \geq 2\  \forall i=1,\ldots, N$, and $1\leq k < N$.

\begin{thm}
\label{thm:param_effi}
For a full-rank weight update matrix, TeRA is more parameter-efficient than VeRA and HiRA, i.e., when $R_i = I_i, \forall i = 1,\ldots,N$, the following holds
\begin{equation}
\sum_{i=1}^{N} R_i \leq J_1+J_2,
\end{equation}
where $\prod_{i=1}^{k} R_i = J_1, \ \prod_{i=k+1}^{N} R_i = J_2$, $R_i \geq 2 \ \forall i=1,\ldots, N$, and $1\leq k < N$.
\end{thm}

\begin{table*}[ht]
\centering
\footnotesize 
\begin{adjustbox}{max width=\textwidth}
\begin{tabular}{llc|cccccccccc}
\toprule
\textbf{Model} & \textbf{Method} & \textbf{Params (\%)} & \textbf{BoolQ} & \textbf{PIQA} & \textbf{SIQA} & \textbf{ARC-c} & \textbf{ARC-e} & \textbf{OBQA} & \textbf{HellaS} & \textbf{WinoG} & \textbf{Average} \\
\midrule
\multirow{10}{*}{Llama-2-7B} 
 & Full FT & 100 & 73.8 & 84.2 & 81.0 & 94.7 & 85.2 & 88.9 & 75.6 & 84.8 & 83.53 \\
\cmidrule(lr){2-12}
 & Prompt Tuning & 0.0012 & 55.93 & 12.35 & 30.50 & 6.06 & 8.63 & 9.40 & 6.91 & 40.57 & 21.29 \\
 & P-Tuning & 0.7428 & 58.75 & 36.02 & 0.20 & 0.17 & 1.98 & 0.80 & 0.01 & 0.00 & 12.24 \\
\cmidrule(lr){2-12}
 
 & LoRA & 0.2484 & \underline{67.65} & 79.22 & 78.20 & 69.20 & 83.88 & 78.60 & 81.05 & 80.98 & 77.35 \\
 & HiRA ($r$=1) & 0.0078 & 65.35 & 77.97 & 72.42 & 62.03 & 81.48 & 64.80 & 79.90 & 70.01 & 71.74 \\
 & HiRA ($r$=32) & 0.2484 & \textbf{69.39} & \textbf{83.24} & \underline{78.86} & \textbf{71.33} & \textbf{86.57} & \textbf{81.40} & \textbf{87.23} & 81.69 & \textbf{79.97} \\
 & MoRA ($r$=32) & 0.2484 & 65.17 & 81.12 & 77.74 & 67.75 & 84.47 & 75.8 & 84.12 & 79.64 & 76.97 \\
 & QuanTA & 0.0408 & 66.88 & \underline{81.28} & 78.45 & \underline{70.82} & \underline{85.40} & 79.4 & 75.86 & \underline{82.16} & 77.53 \\
 & LoRETTA & 0.0052 & 67.52 & 79.33 & 76.31 & 65.70 & 84.34 & 76.00 & \underline{86.32} & 80.98 & 77.06 \\
 & VeRA & \underline{0.0041} & 64.59 & 78.84 & 76.56 & 65.44 & 83.63 & 73.20 & 82.71 & 77.90 & 75.36 \\
 & \textbf{TeRA (Ours)} & \textbf{0.0039} & 65.78 & 81.23 & \textbf{78.92} & 69.45 & 84.05 & \underline{81.00} & 85.94 & \textbf{82.64} & \underline{78.63} \\

\midrule
\multirow{10}{*}{Llama-3-8B} 
 & Full FT & 100 & 75.4 & 88.0 & 81.8 & 96.5 & 89.3 & 93.1 & 83.0 & 86.0 & 86.64 \\
\cmidrule(lr){2-12} 
 & Prompt Tuning & 0.0010 & 56.85 & 45.05 & 36.13 & 31.57 & 32.74 & 29.20 & 14.01 & 50.12 & 36.96 \\
 & P-Tuning & 0.6240 & 59.97 & 11.64 & 8.19 & 7.42 & 8.63 & 9.60 & 1.77 & 37.65 & 18.11 \\
\cmidrule(lr){2-12}
 & LoRA & 0.1695 & \underline{71.99} & 85.91 & 79.58 & 76.19 & 88.55 & 82.60 & 92.54 & 85.63 & 82.88 \\
 & HiRA ($r$=1) & 0.0053 & 68.04 & 85.75 & 75.54 & 76.96 & 90.32 & 77.00 & 89.04 & 77.43 & 80.01 \\
 & HiRA ($r$=32) & 0.1695 & \textbf{73.09} & \underline{88.85} & 81.06 & 80.38 &  \underline{92.68} & \underline{86.20} & 94.37 & \underline{85.87} & \textbf{85.31} \\
 & MoRA ($r$=32) & 0.1692 & 69.05 & 88.25 & 80.14 & 80.89 & \textbf{92.72} & 84.2 & 94.09 & 85.00 & 84.29 \\
 & QuanTA & 0.0343 & 70.03 & 88.41 & \underline{81.67} & \underline{81.05} & 92.21 & 85.6 & \underline{94.68} & 88.47 & \underline{85.27} \\
 & LoRETTA & 0.0045 & 65.17 & \textbf{89.01} & 79.53 & 79.86 & 91.54 & 84.40 & 94.40 & 84.69 & 83.58 \\
 & VeRA & \textbf{0.0022} & 67.95 & 85.64 & 76.51 & 77.22 & 91.29 & 81.00 & 91.96 & 82.24 & 81.73 \\
& \textbf{TeRA (Ours)} & \underline{0.0033} & 70.70 & 88.08 & \textbf{81.58} & \textbf{80.89} & 92.00 & \textbf{88.00} & \textbf{94.92} & \textbf{86.27} & \textbf{85.31} \\
\bottomrule
\end{tabular}
\end{adjustbox}
\caption{Accuracy comparison of different PEFT methods on the Commonsense170k dataset. Full FT performance is cited from \cite{liu2025lift}. The best and second best values among LoRA-style PEFT adapters are highlighted in bold and underlined, respectively.}
\label{tab:common_sense}
\end{table*}

Theorem \ref{thm:param_effi} establishes that TeRA is provably more parameter-efficient than existing methods such VeRA and HiRA when a full-rank weight update matrix is considered. Specifically, TeRA needs at \textit{most} $J_1+J_2$ trainable parameters to parametrize a full-rank update matrix. The number of trainable parameters used in TeRA can also be further reduced by tensorizing the weight matrix to higher-dimensions. For example, a $4096 \times 4096$ matrix can be tensorized to a $64 \times 64 \times 64 \times 64$ tensor. A full-rank update with HiRA or VeRA would require at least $4096 + 4096 = 8192$ parameters, while TeRA requires only $64 + 64 + 64 + 64 = 256$ parameters. By increasing the tensor order $N$ to $24$ (e.g., tensor size of $2 \times 2 \times \cdots \times 2 $), the number of trainable parameters in TeRA can be reduced to as few as $2 \times 24 = 48$.
\begin{thm}
\label{thm:Expressivity}
Consider the optimal weight update $\mathbf{W}^\star \in \mathbb{R}^{J_1 \times J_2}$ and the TeRA weight update, $\mathbf{W}_{TeRA} \in \mathbb{R}^{J_1 \times J_2}$, whose tensorized format is defined in Eq. (\ref{eq:our_method}). Denote $\bigotimes_{i=1}^{k}\mathbf{A}^{(i)}$ by $\mathbf{L}^{\top} \in \mathbb{R}^{\prod_{i=1}^k R_i \times J_1}$, $\bigotimes_{i=k+1}^{N}\mathbf{A}^{(i)}$ by $\mathbf{M} \in \mathbb{R}^{J_2 \times \prod_{i=k+1}^N R_i}$, and $\mathbf{Z} = \mathbf{L}^{\dagger} \mathbf{\mathbf{W}^\star} \mathbf{M}^{\dagger} \oslash \mathbf{G}_{[N;k]}$, where $\oslash$ is the element-wise division. Then, the following inequality holds
\begin{equation}
\begin{aligned}
\label{eq:expressivity}
& \min_{\{\operatorname{diag}(\mathbf{d}^{(i)}) \}_{i=1}^N} \| \mathbf{\mathbf{W}^\star} - \mathbf{W}_{TeRA } \|_F^2 \\
&  \leq \|\mathbf{\mathbf{W}^\star}- \mathbf{L}\mathbf{L}^{\dagger}\mathbf{\mathbf{W}^\star}\mathbf{M}^{\dagger}\mathbf{M}\|_F^2 \\
& + g_{max}^2 (\|\mathbf{Z}\|_F^2 - \|\operatorname{Fold}_{[N;k]}(\mathbf{Z})\|_2^2) \|\mathbf{L}\|_F^2\|\mathbf{M}\|_F^2,
\end{aligned}
\end{equation}
where $g_{max}$ is the largest entry in $\mathcal{G}$,  and $\|\cdot\|_2$ denotes the tensor spectral norm.
\end{thm} 

Theorem \ref{thm:Expressivity} provides a theoretical upper bound on the approximation error between the optimal weight update and the TeRA weight update. This bound consists of two terms. The first term, $\|\mathbf{\mathbf{W}^\star}- \mathbf{L}\mathbf{L}^{\dagger}\mathbf{\mathbf{W}^\star}\mathbf{M}^{\dagger}\mathbf{M}\|_F^2$, quantifies the portion of the optimal update $\mathbf{W}^\star$, which lies outside the subspace characterized by the frozen factor matrices $\{\mathbf{A}^{(i)}\}_{i=1}^N$. This error can only be minimized by expanding the subspace through increasing the ranks, $\{R_i\}_{i=1}^N$. Consequently, the bound provides direct theoretical motivation to maximize the ranks to their corresponding tensor dimension sizes, \textit{i.e.}, $R_i = I_i, \ \forall i = 1,2,\dots,N$. As a side benefit, this also reduces the number of hyperparameters in TeRA by eliminating the need to choose the tensor network ranks $[R_1, R_2, \ldots, R_N] $.

The second error term establishes the trade-off between parameter efficiency and approximation accuracy (expressivity) in TeRA. It is bounded by a term dependent on the tensor spectral norm of $\mathbf{Z}$~\cite{de2000best}. The tensor spectral norm has been shown to decrease as the order of the tensor, $N$, increases~\cite{wang2017operator}. Therefore, assuming $R_i = I_i \  \forall i=1,\ldots, N$, this indicates that using a higher-order tensorization (a larger $N$ in $\Delta\mathcal{W} \in \mathbb{R}^{I_1 \times I_2 \times \cdots \times I_N}$) reduces the number of trainable parameters ($\sum_i^N R_i$), but enlarges the upper bound of the approximation error. Conversely, lower-order tensorization may result in a tighter error bound, but at the cost of more trainable parameters. The proofs for the three theorems are provided in the appendix.

\begin{table*}[ht]
\centering
\footnotesize 
\begin{adjustbox}{max width=\textwidth}
\begin{tabular}{llcccccccc}
\toprule
\textbf{Model} & \textbf{Method} & \textbf{Params (\%)} & \textbf{BLEU} & \textbf{BERT F1} & \textbf{BERT-R} & \textbf{BERT-P} & \textbf{Meteor} & \textbf{R-L} & \textbf{Average} \\
\midrule
\multirow{8}{*}{Llama-2-7B} 
& Prompt Tuning & 0.0012 & 0.04 & 72.44 & 77.38 & 68.23 & 0.80 & 0.80 & 36.62 \\
& P-Tuning & 0.7428 & 0.60 & 83.29 & 83.33 & 83.28 &15.11 & 12.36 & 46.33 \\
\cmidrule(lr){2-10}
& LoRA & 0.2484 & 2.59 & 85.25 & 85.31 & \textbf{85.21} & \underline{14.14} & \underline{13.28} & \underline{47.63} \\
& HiRA ($r=1$) & 0.0078 & 2.39 & 84.84 & 84.93 & 84.78 & 13.35 & 12.60 & 47.15  \\
& HiRA ($r=32$) & 0.2484 & \textbf{2.67} & \underline{85.26} & \underline{85.37} & 85.17 & 13.94 & 13.19 & 47.60 \\
& MoRA ($r=32$) & 0.2484 & 1.91 & 84.85 & 84.91 & 84.83 & 11.80 & 11.59 & 46.65 \\
& QuanTA & 0.0408 & 2.65 & 85.14 & 85.32 & 85.00 & 14.19 & 13.32 & 47.60 \\
& LoRETTA & 0.0052 & 2.22 & 84.99 & 84.97 & 85.03 & 12.37 & 11.85 & 46.91 \\
& VeRA & \underline{0.0041} & 2.27 & 85.01 & 85.04 & 85.02 & 12.64 & 12.55 & 47.09 \\
& \textbf{TeRA (Ours)} & \textbf{0.0039} & \underline{2.62} & \textbf{85.28} & \textbf{85.40} & \underline{85.19} & \textbf{14.26} & \textbf{13.39} & \textbf{47.69} \\
\midrule
\multirow{8}{*}{Llama-3-8B} 
& Prompt Tuning & 0.0010 & 1.45 & 82.99 & 82.99 & 83.05 & 14.72 & 13.13 & 46.39 \\
& P-Tuning & 0.6240 & 1.50 & 81.52 & 81.07 & 82.01 & 15.49 & 13.55 & 45.86 \\
\cmidrule(lr){2-10}
& LoRA & 0.1695 & 3.24 & \underline{85.02} & \underline{84.49} & \textbf{85.60} & \underline{15.16} & 14.14 & \underline{47.94} \\
& HiRA ($r=1$) & 0.0053 & \underline{3.36} & 84.81 & 84.40 & 85.26 & 15.12 & \underline{14.19} & 47.86 \\
& HiRA ($r=32$) & 0.1695 & 3.22 & 84.58 & 84.35 & 84.87 & 14.90 & 13.72 & 47.61 \\
& MoRA ($r=32$) & 0.1692 & 2.35 & 84.20 & 83.79 & 84.67 & 11.96 & 11.51 & 46.41 \\
& QuanTA & 0.0343 & 3.04 & 84.55 & 84.19 & 84.97 & 13.90 & 13.08 & 47.29 \\
& LoRETTA & 0.0045 & 2.97 & 84.77 & 84.26 & 85.31 & 13.78 & 13.15 & 47.37 \\
& VeRA & \underline{0.0022} & 3.12 & 84.61 & 84.27 & 85.01 & 14.56 & 13.72 & 47.55 \\
& \textbf{TeRA (Ours)} & \textbf{0.0021} & \textbf{3.38} & \textbf{85.03} & \textbf{84.53} & \underline{85.57} & \textbf{15.32} & \textbf{14.59} & \textbf{48.07} \\
\bottomrule
\end{tabular}
\end{adjustbox}
\caption{Evaluation results of different PEFT methods on the ConvAI2 dataset. The considered metrics include BLEU, BERTScore (F1/R/P), Meteor, and ROUGE-L.}
\label{tab:convai2}
\end{table*}

\section{Experiments}
We conducted extensive experiments to demonstrate the effectiveness of TeRA across a diverse set of reasoning and generation tasks in English. We also performed a series of ablation studies to validate our model design and analyze the impact of hyperparameter choices.

\paragraph{Implementation Details.}  We used two LLMs as the base models for fine-tuning: Llama-2-7B and Llama-3-8B. The percentage of trainable parameters was calculated as $\frac{\text{\# trainable params}}{\text{\# total params}}$, where \textit{trainable parameters} refers to those requiring gradient updates, and \textit{total parameters} include both the frozen and trainable parameters across all layers in the LLM. In practice, we find that tensorizing only one dimension of the weight update matrix in TeRA yields a good trade-off between performance and the number of trainable parameters. To reduce the hyperparameters search space, we always tensorize each dimension into equal-sized modes and set $R_i = I_i \ \forall i=1,\ldots, N$. E.g., a dimension of size $4096$ can be tensorized into $64\times 64$, $16\times 16\times 16\times 16$, etc. We report the average performance of TeRA over 5 independent runs.

\paragraph{Baseline Methods.} We benchmark TeRA against two main categories of PEFTs: prompt-based methods (Prompt-Tuning, P-Tuning) and LoRA-style adapters with no inference overhead (LoRA, HiRA, MoRA, QuanTA, LoRETTA, VeRA). To ensure fairness in terms of number of trainable parameters, following \cite{kopiczko2024vera}, we applied all methods to the query and value weights in the attention modules. We also report HiRA with two rank settings, $r\in \{1,32\}$, to show how it performs with different number of trainable parameters.

\subsection{Commonsense Reasoning} 
Following \citet{huang2025hira}, we evaluated TeRA on eight commonsense reasoning tasks using the Commonsense170k benchmark \cite{hu2023llm}, which has $170,420$ query-answer pairs. The eight sub-tasks include: BoolQ \cite{clark2019boolq}, PIQA \cite{bisk2020piqa}, SIQA \cite{sap2019socialiqa}, HellaSwag \cite{zellers2019hellaswag}, WinoGrande \cite{sakaguchi2020winogrande}, ARC-e and ARC-c \cite{clark2018think}, and OBQA \cite{mihaylov2018can}. The test accuracy is given in Table \ref{tab:common_sense}.

\paragraph{Results.} Table \ref{tab:common_sense} reveals that TeRA consistently outperforms baseline methods which require similar number of trainable parameters, such as HiRA ($r=1$), LoRETTA, and VeRA, in terms of average accuracy. Specifically, TeRA achieved an average accuracy of $78.63\%$ on Llama-2-7B (followed by $77.06\%$ with LoRETTA) and $85.31\%$ on Llama-3-8B (followed by $83.58\%$ with LoRETTA). TeRA matched the performance of the best performing high-rank adapter, HiRA ($r=32$), while having $64\times$ fewer trainable parameters in the Llama-2-7B model and $51\times$ less trainable parameters in the Llama-3-8B model. Remarkably, TeRA requires fewer parameters than even the lowest-rank HiRA configuration ($r=1$), yet consistently delivering superior performance. These results demonstrate that TeRA achieves the performance benefits of high-rank adapters, while offering significantly improved parameter efficiency.

\paragraph{High-Rank Weight Updates of TeRA.} To visualize the high-rank nature of TeRA, Figure \ref{fig:rank_different_methods} shows the ranks of the update matrices across different layers obtained in the commonsense reasoning task for Llama-3-8B. Observe that TeRA consistently obtained high-rank updates. Additionally, the weight updates of TeRA often reach full-rank, validating Theorem \ref{thm:full-rank} empirically. In
comparison, the low-rank weight updates of methods such
as LoRA, LoRETTA, and VeRA may limit their expressivity. 

\subsection{Personalized Dialogue Generation} 

We evaluated the ability of fine-tuned models to engage in natural conversations using the ConvAI2 dataset \cite{dinan2019second} with 17, 878 training and 1, 000
testing multi-turn conversations. Following the experimental setup of \cite{huang2025hira}, where only the speaker’s persona is revealed (self-persona setting), we report the quality of the generated responses using standard metrics, including BLEU, BERTScore \cite{Zhang2020BERTScore}, METEOR \cite{banerjee2005meteor}, and ROUGE \cite{lin2004rouge} in Table \ref{tab:convai2}.

\paragraph{Results.} Table \ref{tab:convai2} shows that TeRA consistently achieved the highest average score among all baseline methods in conversational tasks and used the least number of trainable parameters among LoRA-style adapters. More specifically, TeRA achieved an average score of $47.69\%$ with Llama-2-7B and $48.07\%$ with Llama-3-8B. This shows the superior performance and parameter efficiency of TeRA in conversation fine-tuning tasks.

\subsection{Arithmetic Reasoning}
We evaluated arithmetic reasoning capabilities of the fine-tuned model through the Math10k dataset from \cite{cobbe2021training, koncelkedziorskietal2016mawps, ling2017program}, consisting of 10, 000 mathematical reasoning examples. We considered two sub-tasks: AQuA \cite{ling2017program} and SVAMP \cite{pateletal2021nlp}. The test accuracies are reported in Table \ref{tab:maths}.

\begin{table}[t]
\centering
\footnotesize
\setlength{\tabcolsep}{3pt}
\begin{tabular}{@{}llcccc@{}}
\toprule
\textbf{Model} & \textbf{Method} & \textbf{Params (\%)} & \textbf{AQuA} & \textbf{SVAMP} \\
\midrule
\multirow{5}{*}{%
  \begin{tabular}{@{}c@{}}
    Llama\\
    -2-7B
  \end{tabular}}
& HiRA ($r=1$) & 0.0078 & \underline{18.90} & 27.40 \\
& HiRA ($r=32$) & 0.2484 & \underline{18.90} & 46.50 \\
& LoRETTA  & 0.0052 & 12.60 & \underline{47.50} \\
& VeRA & \underline{0.0041} & 18.50 & 26.60 \\
& \textbf{TeRA (Ours)} & \textbf{0.0039} & \textbf{24.41} & \textbf{49.70} \\
\midrule
\multirow{5}{*}{%
  \begin{tabular}{@{}c@{}}
    Llama\\
    -3-8B
  \end{tabular}}
& HiRA ($r=1$) & 0.0053 & 27.95 & 57.70 \\
& HiRA ($r=32$) & 0.1695 & \underline{29.92} & \underline{72.80}\\
& LoRETTA & 0.0045 & 18.50 & 56.50 \\
& VeRA & \underline{0.0022} & 25.59 & 55.70 \\
& \textbf{TeRA (Ours)} & \textbf{0.0021} & \textbf{30.71} & \textbf{73.10} \\
\bottomrule
\end{tabular}
\caption{Performance comparison in terms of test accuracy on arithmetic reasoning datasets.}
\label{tab:maths}
\end{table}

\paragraph{Results.} TeRA consistently outperformed all baseline methods in AQuA and SVAMP while requiring the least number of trainable parameters. As shown in Table \ref{tab:maths}, with the Llama–2‑7B backbone, TeRA achieved accuracies of \(24.41\%\) on AQuA and \(49.7\%\) on SVAMP, thus outperforming the strongest baseline by \(5.51\%\) and \(1.7\%\) absolute points, respectively. With Llama–3‑8B, TeRA obtains \(30.71\%\) on AQuA and \(73.1\%\) on SVAMP, outperforming HiRA (\(r=32\)) by \(0.79\%\) and \(0.3\%\) absolute points while using \(80\times\) fewer parameters.

\subsection{Ablation study}

\begin{figure}[ht]
    \centering
    \includegraphics[width=0.48\textwidth]{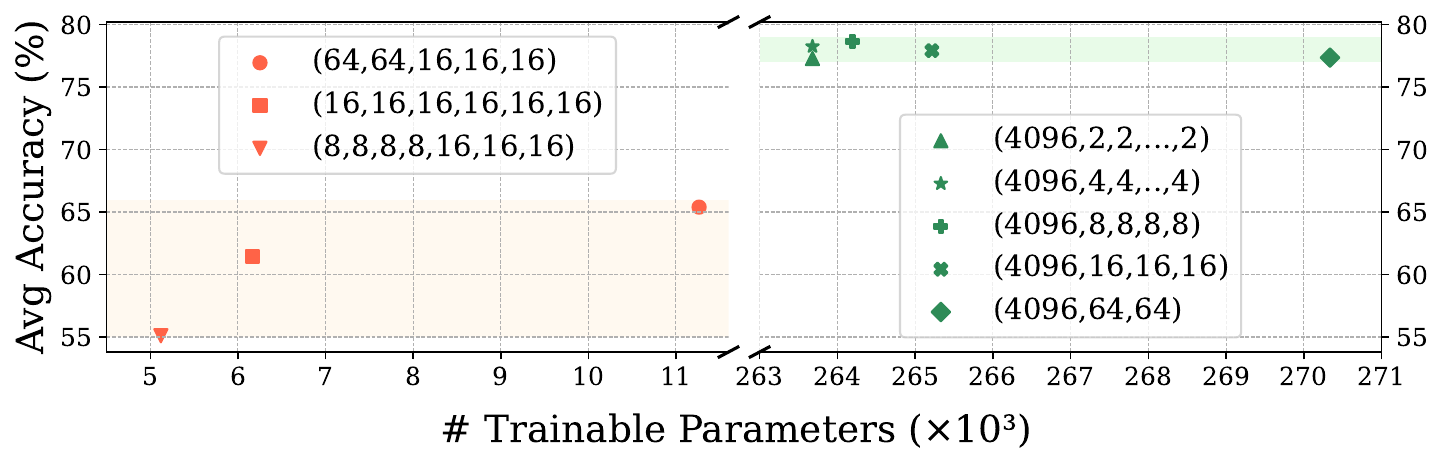}
    \caption{Average accuracies across eight commonsense reasoning tasks against number of trainable parameters under different tensorization strategies in Llama-2-7B. } 
    \label{fig:tensorization_on_performance}
\end{figure}

\paragraph{Impact of Tensorization on Performances.} Different tensorizations of the weight matrices lead to different trade-offs between the number of trainable parameters and the model performance, as formalized in Theorem \ref{thm:Expressivity}. Figure \ref{fig:tensorization_on_performance} (left) shows that when both dimensions of the original weight matrix are tensorized, the performance decreases rapidly with decrease in the number of trainable parameters. Empirically, we achieve an optimal trade-off between performance and parameter efficiency by tensorizing only one dimension of the original weight matrix. Under this approach, the performance of TeRA remains robust across different tensorization sizes (See Figure \ref{fig:tensorization_on_performance} (right)).

\begin{figure}[ht]
    \centering
    \includegraphics[width=0.48\textwidth]{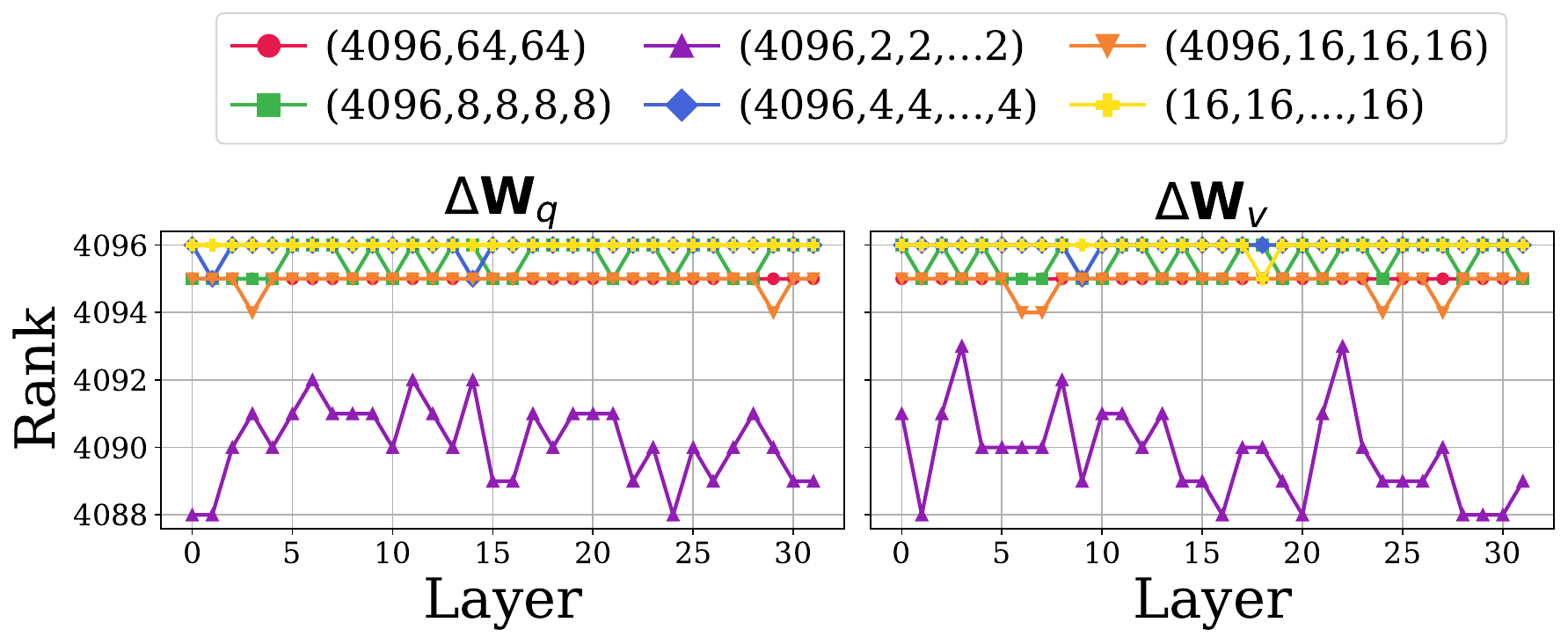}
    \caption{Ranks of $\Delta \mathbf{W}_q$ and $\Delta \mathbf{W}_v$ (Largest possible rank = 4096) in Llama-2-7B under different tensorization schemes in the commonsense reasoning task. } 
    \label{fig:rank_ours_different_splits}
\end{figure}

\paragraph{Impact of Tensorization on Ranks.} The high-rank property of TeRA is insensitive to the specific choice of tensorization schemes, as formalized in Theorem \ref{thm:full-rank}. We evaluated the ranks of the query weight updates, $\Delta \mathbf{W}_q$, and the value weight updates, $\Delta \mathbf{W}_v$, across different layers under various tensorizations. Figure~\ref{fig:rank_ours_different_splits} shows that, as desired, TeRA obtains high-rank (near full-rank) updates across different tensorization choices.

\begin{figure}[ht]
    \centering
    \includegraphics[width=0.42\textwidth]{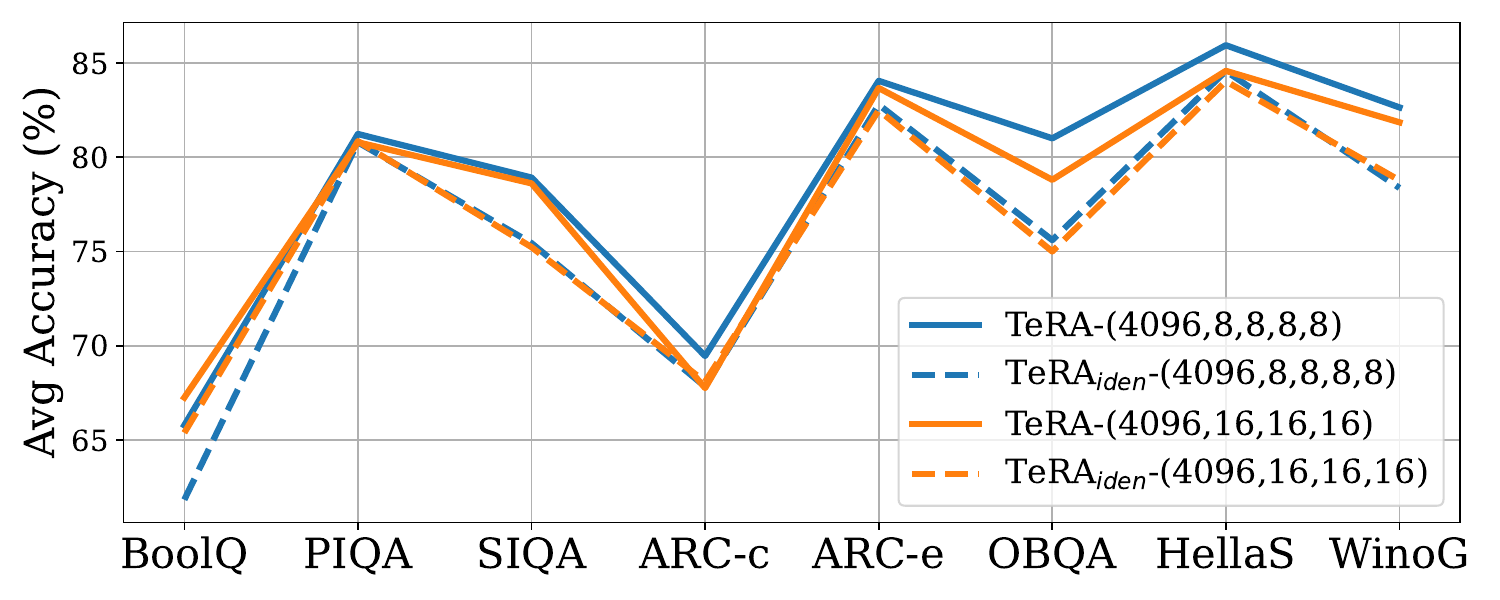}
    \caption{Comparison between TeRA and TeRA$_{iden}$ on the commonsense reasoning dataset with Llama-2-7B. } 
    \label{fig:tera_vs_tera_iden}
\end{figure}

\paragraph{Initialization of Frozen Factor Matrices.} To explore different initialization choices for the frozen factor matrices, we compared TeRA with its variant, $\text{TeRA}_{iden}$, where its frozen factor matrices are all identity matrices. Note that $\text{TeRA}_{iden}$ has the same number of trainable parameters as TeRA. Figure \ref{fig:tera_vs_tera_iden} shows that,  with identical hyperparameters, TeRA consistently outperforms $\text{TeRA}_{iden}$ in terms of average accuracy, highlighting the effectiveness of the random tensor network initialization scheme in TeRA. We also experimented with non-sharing factor tensors and report the result in Appendix \ref{sec:non-sharing}.

\section{Conclusion}
We have introduced TeRA, a high-rank PEFT adapter which employs a tensor network approach to parametrize the tensorized weight updates. Such a parametrization has allowed TeRA to offer a more effective alternative to existing vector-based adapters with a similar amount of trainable parameters, while achieving much better performances and high-rank updates.  TeRA is particularly well-suited for large-scale customization scenarios, where hundreds of thousands of task- or user-specific adapters may be required. Its ability to maintain high expressivity while using extremely few trainable parameters, and requiring minimal storage per adapter, makes it an attractive solution for scalable personalization of LLMs.

\section{Limitations}
Despite the good performance of TeRA and its low number of trainable parameters, the training process can be time-consuming due to the tensor contractions required for fine-tuning Large Language Models (LLMs). A promising direction for future work is the integration of TeRA with custom hardware specifically designed for accelerating these tensor operations. Furthermore, TeRA currently does not employ low-bit quantization methods. Incorporating such techniques could further reduce the memory footprint, making the method more accessible in resource-constrained settings \cite{zhou2023dybit}. 

Moreover, although Theorem \ref{thm:Expressivity} has provided the theoretical foundation for significantly minimizing the search space of selecting the best tensorization scheme, identifying the optimal configuration for a given task still involves sampling different settings. Future work could focus on developing efficient methods for the selection of tensorization schemes for compressing neural network weights. We believe that our work can pave the way for future explorations in these avenues.

\bibliography{custom}

\newpage
\clearpage
\appendix
\label{sec:appendix}

\section{Preliminaries}
The mathematical notations used in Appendix are listed in Table \ref{tab:math_notation_appendix}. This is consistent with the notation used in \citet{7038247}. See also the Preliminaries section of the main paper.

\begin{table}[htbp]
\footnotesize

  \centering
  \begin{tabular}{cc}
    \toprule
      Symbol   & Meaning \\
    \midrule
    $a$, $\mathbf{a}$, $\mathbf{A}$, $\mathcal{A}$    & Scalar, vector, matrix, tensor  \\
    $(\cdot)^\top$ & Matrix transpose \\
    $(\cdot)^{\dagger}$ & Matrix pseudoinverse \\
    $\|\cdot\|_F$ & Frobenius norm \\
    $\mathcal{A}(i_1, i_2, \dots, i_N)$ &  The $(i_1, \dots, i_N)$-th element of $\mathcal{A}$ \\
    $\mathbf{a} \circ \mathbf{b}$ & Outer product between two vectors \\
    $\mathbf{A} \otimes \mathbf{B}$ & Kronecker product \\
    $\mathcal{A} \times_n \mathbf{B}$ & Mode-$n$ product\\
    diag$\left(\mathbf{a}\right)$ & Diagonal matrix with diagonal as $\mathbf{a}$ \\
    $\text{Fold}_{[N;k]}(\mathbf{A}_{[N;k]})$ & Fold matrix $\mathbf{A}_{[N;k]}$ to tensor $\mathcal{A}$\\
    
    \bottomrule
  \end{tabular}
    \caption{Mathematical notations}
  \label{tab:math_notation_appendix}
\end{table}

\section{Theorem 1 and Its Proof}
\paragraph{TeRA Formulation.} We parametrize the weight update tensor $\Delta \mathcal{W}$ as the mode-$n$ product of a frozen core tensor $\mathcal{G}\in \mathbb{R}^{R_1 \times R_2 \times \cdots \times R_N}$, a set of $N$ frozen non-diagonal factor matrices $\{\mathbf{A}^{(i)}\in \mathbb{R}^{R_i \times I_i}\}_{i=1}^{N}$, and a set of $N$ trainable vectors $\{\mathbf{d}^{(i)} \in \mathbb{R}^{R_i}\}_{i=1}^N$, which are diagonal entries of $\{\operatorname{diag}(\mathbf{d}^{(i)})\in \mathbb{R}^{R_i \times R_i}\}_{i=1}^N$. The definition of TeRA is 
\begin{equation}
\begin{aligned}
\label{eq:our_method_appendix}
\Delta \mathcal{W} = \mathcal{G} &\times_1 \operatorname{diag}(\mathbf{d}^{(1)}) \times_2 \operatorname{diag}(\mathbf{d}^{(2)}) \times_3 \\
& \cdots \times_N \operatorname{diag}(\mathbf{d}^{(N)}) \times_1 \mathbf{A}^{(1)}\\
& \times_2 \mathbf{A}^{(2)} \times_3 \cdots \times_N \mathbf{A}^{(N)}.
\end{aligned} 
\end{equation}
or equivalently
\begin{equation}
\begin{aligned}
&\Delta \mathcal{W}(i_1,\ldots,i_N) \\
&= \sum_{r_1=1}^{R_1} \sum_{r_2=1}^{R_2} \cdots \sum_{r_N=1}^{R_N}\mathcal{G}(r_1,r_2,\ldots,r_N) \\
&\quad \quad \mathbf{d}^{(1)}(r_1) \mathbf{d}^{(2)}(r_2)\cdots \mathbf{d}^{(N)}(r_N)\\
& \quad \quad\mathbf{A}^{(1)}(r_1,i_1) \mathbf{A}^{(2)}(r_2,i_2) \cdots \mathbf{A}^{(N)}(r_N,i_N)
\end{aligned} 
\end{equation}

\setcounter{thm}{0}
\begin{thm}
Let $\Delta \mathbf{W} \in \mathbb{R}^{J_1 \times J_2}$ be the weight update matrix, and $\Delta \mathcal{W} \in \mathbb{R}^{I_1 \times I_2 \times \cdots \times I_N} = \operatorname{Fold}_{[N;k]}(\Delta \mathbf{W}_{[N;k]} )$ be its folded weight update tensor, parametrized by TeRA as in Eq. (\ref{eq:our_method_appendix}). The following inequality holds
\begin{equation}
\operatorname{rank}\left(\Delta \mathbf{W}_{[N;k]}\right) \leq \operatorname{min}\left(\prod_{i=1}^k R_i, \prod_{i=k+1}^N R_i\right).
\end{equation} 
This allows for a full-rank update matrix under any tensorization (folding) schemes if $R_i = I_i \ \forall i = 1,\ldots,N$, i.e., $\operatorname{rank}\left(\Delta \mathbf{W}_{[N;k]}\right)  \leq \operatorname{min}\left(J_1, J_2\right)$ .
\end{thm}

\begin{proof}
We first group the diagonal and non‑diagonal factor matrices that correspond to the same mode and rewrite Eq. (\ref{eq:our_method_appendix}) as a standard Tucker decomposition \cite{tucker1964extension}
\begin{equation}
\label{eq:Tucker_b}
\Delta\mathcal{W} =\mathcal{G}\times_1\mathbf{B}^{(1)}\times_2\mathbf{B}^{(2)}\cdots\times_N\mathbf{B}^{(N)}, 
\end{equation}
where $\mathbf{B}^{(i)} := \operatorname{diag}(\mathbf{d}^{(i)})\mathbf{A}^{(i)} \in\mathbb{R}^{R_i\times I_i}$. Since $\mathbf{A}^{(i)}$ for $i=1,\ldots, N$ are randomly initialized using Kaiming initialization \cite{he2015delving}, they are usually full-rank, making $\mathbf{B}^{(i)}$ also full-rank given no zero entries in $\mathbf{d}^{(i)}$.

Recall that the tensorization was chosen such that
\begin{equation}
J_1=\prod_{i=1}^{k} I_i,
\qquad
J_2=\prod_{i=k+1}^{N} I_i.
\end{equation}

Using the Kronecker product, Eq. (\ref{eq:Tucker_b}) can be written in a matrix format as 
\begin{equation}
\Delta\mathbf{W}_{[N;k]}
=\left(\bigotimes_{i=1}^{\,k}\mathbf{B}^{(i)}\right)^{\top}\mathbf{G}_{[N;k]}\left(\bigotimes_{i=k+1}^{\,N}\mathbf{B}^{(i)}\right),
\end{equation}
where $\otimes$ denotes the Kronecker product, and $\mathbf{G}_{(1:N,k)} \in\mathbb{R}^{\left(\prod_{i=1}^{k}R_i\right)\times \left(\prod_{i=k+1}^{N}R_i\right)}$
is a matricization of the core tensor.

Therefore, the following holds
\begin{equation}
\begin{aligned}
\operatorname{rank} & \left(\Delta \mathbf{W}_{[N;k]}\right) \le \min \Biggl\{ \operatorname{rank}\mathbf{\left(\bigotimes_{i=1}^{\,k}\mathbf{B}^{(i)}\right)^{\top}}, \\
&\operatorname{rank}\mathbf{\left(\mathbf{G}_{[N;k]}\right)},\operatorname{rank}\mathbf{\left(\bigotimes_{i=k+1}^{N}\mathbf{B}^{(i)}\right)} \Biggr\}.
\end{aligned}
\end{equation}
As ranks, $R_i$, are set to be smaller or equal to than their corresponding mode sizes, $I_i$, $\prod_{i=1}^{k}I_i \geq \prod_{i=1}^{k}R_i$ and $\prod_{i=k+1}^{N}I_i\geq \prod_{i=k+1}^{N}R_i$. Therefore,
\begin{equation}
  \operatorname{rank}(\Delta\mathbf{W}_{[N;k]}) \le  \min \Bigl(\prod_{i=1}^{k}R_i,\ \prod_{i=k+1}^{N}R_i \Bigr). 
\end{equation}
Thus, if $I_i = R_i$ for $i = 1,\ldots,N$
\begin{equation}
\begin{aligned}
    \operatorname{rank}\left(\Delta \mathbf{W}_{[N;k]}\right) &\leq \min \Bigl(\prod_{i=1}^{k}I_i,\ \prod_{i=k+1}^{N}I_i \Bigr) \\
    &= \operatorname{min}\left(J_1, J_2\right).
\end{aligned}
\end{equation}

\end{proof}

\section{Theorem 2 and Its Proof}

\begin{thm}
For a full-rank weight update matrix, TeRA is more parameter-efficient than VeRA and HiRA, i.e., when $R_i = I_i, \forall i = 1,\ldots,N$, the following holds
\begin{equation}
\sum_{i=1}^{N} R_i \leq J_1+J_2,
\end{equation}
where $\prod_{i=1}^{k} R_i = J_1, \ \prod_{i=k+1}^{N} R_i = J_2$, $R_i \geq 2 \ \forall i=1,\ldots, N$, and $1\leq k < N$.
\end{thm}

\begin{proof}
Since $R_i = I_i, \forall i = 1,\ldots,N$, we just need to prove that $I_1 + I_2 + \cdots + I_k \leq J_1$ for $I_1 \times I_2 \times \cdots \times I_k = J_1$. Then, we naturally have $I_1 + I_2 + \cdots + I_N \leq J_1+J_2$, such that $I_1 \times I_2 \times \cdots \times I_k = J_1, \ I_{k+1} \times I_{k+2} \times \cdots \times I_N = J_2$, $\{I_i \geq 2\}_{i=1}^{N}$, and $1\leq k < N$.

When $N = k = 1$, $I_1 = J_1$, thus $I_1 = J_1$. The induction hypothesis is such that assume for some $N = k \geq 1$, we have $\sum_{i=1}^{k} I_i \leq \prod_{i=1}^{k} I_i$. For  $a, b\geq 2$, $(a-1)(b-1) \geq 1$. Therefore, $a \cdot b \geq a + b$. This in turn yields for $N=k+1 \geq 1$, $I_{k+1} + \sum_{i=1}^{k} I_i  \le I_{k+1} \cdot \sum_{i=1}^{k} I_i $, as $\sum_{i=1}^{k} I_i, I_{k+1} \geq 2$. Therefore, we have proven that $\sum_{i=1}^{k+1} I_i \leq \prod_{i=1}^{k+1} I_i$.  As this holds for $k+1$, by mathematical induction, the following holds for $k\geq1$, $I_1 + I_2 + \cdots + I_k \leq J_1$ where $I_1 \times I_2 \times \cdots \times I_k = J_1$. Applying this result twice, we have $I_1 + I_2 + \cdots + I_N \leq J_1+J_2$. Therefore, $\sum_{i=1}^{N} R_i \leq J_1+J_2$.

\end{proof}
    
\section{Theorem 3 and Its Proof}
\begin{table*}[!ht]
\centering
\begin{adjustbox}{max width=\textwidth}
\begin{tabular}{lccccccccccc}
\toprule
\textbf{Target Modules} & \textbf{Params (\%)} & \textbf{BoolQ} & \textbf{PIQA} & \textbf{SIQA} & \textbf{ARC-c} & \textbf{ARC-e} & \textbf{OBQA} & \textbf{HellaS} & \textbf{WinoG} & \textbf{Average} \\
\midrule
Query only & 0.0020 & 65.54 & 78.13 & 72.06 & 60.07 & 78.62 & 67.4 & 79.62 & 75.06 & 72.06 \\
Value only & 0.0020 & 62.94 & 80.36 & 77.69 & 67.75 & 82.49 & 78.2 & 84.52 & 81.61 & 76.94 \\
Query \& Value & 0.0040 & \textbf{65.78} & \textbf{81.23} & \textbf{78.92} & \textbf{69.45} & \textbf{84.05} & \textbf{81.00} & \textbf{85.94} & \textbf{82.64} & \textbf{78.63} \\
\bottomrule
\end{tabular}
\end{adjustbox}
\caption{The difference between applying TeRA only on query matrices, value matrices, and both query and value matrices in Llama-2-7B on the commonsense reasoning dataset. The tensorization scheme is $(4096, 8, 8, 8, 8)$.}
\label{tab:q,v,qv}
\end{table*}
\begin{thm}
Consider the optimal weight update $\mathbf{W}^\star \in \mathbb{R}^{J_1 \times J_2}$ and the TeRA weight update, $\mathbf{W}_{TeRA} \in \mathbb{R}^{J_1 \times J_2}$, whose tensorized format is defined in Eq. (\ref{eq:our_method}). Denote $\bigotimes_{i=1}^{k}\mathbf{A}^{(i)}$ by $\mathbf{L}^{\top} \in \mathbb{R}^{\prod_{i=1}^k R_i \times J_1}$, $\bigotimes_{i=k+1}^{N}\mathbf{A}^{(i)}$ by $\mathbf{M} \in \mathbb{R}^{J_2 \times \prod_{i=k+1}^N R_i}$, and $\mathbf{Z} = \mathbf{L}^{\dagger} \mathbf{\mathbf{W}^\star} \mathbf{M}^{\dagger} \oslash \mathbf{G}_{[N;k]}$, where $\oslash$ is the element-wise division. Then, the following inequality holds
\begin{equation}
\begin{aligned}
\label{eq:expressivity_appendix}
& \min_{\{\operatorname{diag}(\mathbf{d}^{(i)}) \}_{i=1}^N} \| \mathbf{\mathbf{W}^\star} - \mathbf{W}_{TeRA } \|_F^2 \\
&  \leq \|\mathbf{\mathbf{W}^\star}- \mathbf{L}\mathbf{L}^{\dagger}\mathbf{\mathbf{W}^\star}\mathbf{M}^{\dagger}\mathbf{M}\|_F^2 \\
 &  + g_{max}^2(\|\mathbf{Z}\|_F^2 - \|\operatorname{Fold}_{[N;k]}(\mathbf{Z})\|_2^2) \|\mathbf{L}\|_F^2\|\mathbf{M}\|_F^2,
\end{aligned}
\end{equation}
where $g_{max}$ is the largest entry in $\mathcal{G}$,  and $\|\cdot\|_2$ denotes the tensor spectral norm.
\end{thm} 

\begin{proof}
In the analysis, we assume that there is no zero in $\mathbf{G}_{[N;k]}$. Let $\mathbf{W}^\star \in \mathbb{R}^{J_1\times J_2}$ be the optimal weight update and $\mathbf{W}_{TeRA}$ be the learned TeRA update. Denote $\bigotimes_{i=1}^{k}\mathbf{A}^{(i)}$ by $\mathbf{L}^{\top} \in \mathbb{R}^{\prod_{i=1}^k R_i \times J_1}$, and $\bigotimes_{i=k+1}^{N}\mathbf{A}^{(i)}$ by $\mathbf{M} \in \mathbb{R}^{J_2 \times \prod_{i=k+1}^N R_i}$.

Using Kronecker product, Eq. (\ref{eq:Tucker_b}) can be written in matrix format as 
\begin{equation}
\begin{aligned}
&\mathbf{W}_{TeRA} = \left(\bigotimes_{i=1}^{\,k}\operatorname{diag}(\mathbf{d}^{(i)})\mathbf{A}^{(i)}\right)^{\top} \mathbf{G}_{[N;k]} \\
&\left(\bigotimes_{i=k+1}^{\,N}\operatorname{diag}(\mathbf{d}^{(i)})\mathbf{A}^{(i)}\right),
\end{aligned}
\end{equation}

Let $\mathbf{E} = \bigotimes_{i=1}^{\,k}\operatorname{diag}(\mathbf{d}^{(i)})$ and $\mathbf{F} = \bigotimes_{i=k+1}^{\,N}\operatorname{diag}(\mathbf{d}^{(i)})$. Using the Kronecker identity, we can write 
\begin{equation}
\mathbf{W}_{TeRA}
=\mathbf{L} \mathbf{E} \mathbf{G}_{[N;k]}\mathbf{F}\mathbf{M},
\end{equation}

Thus, our aim is to minimize the following discrepancy measured in the Frobenius norm
\begin{equation}
\| \mathbf{W}^\star - \mathbf{L} \mathbf{E} \mathbf{G}_{[N;k]}\mathbf{F}\mathbf{M} \|_F^2.
\end{equation}

As $\mathbf{W}^\star - \mathbf{L} \mathbf{E} \mathbf{G}_{[N;k]}\mathbf{F}\mathbf{M} = \mathbf{W}^\star -\mathbf{L}\mathbf{L}^{\dagger}\mathbf{\mathbf{W}^\star}\mathbf{M}^{\dagger}\mathbf{M} + \mathbf{L}\mathbf{L}^{\dagger}\mathbf{\mathbf{W}^\star}\mathbf{M}^{\dagger}\mathbf{M} - \mathbf{L} \mathbf{E} \mathbf{G}_{[N;k]}\mathbf{F}\mathbf{M} $, we have

\begin{equation}
\begin{aligned}
&\| \mathbf{W}^\star - \mathbf{L} \mathbf{E} \mathbf{G}_{[N;k]}\mathbf{F}\mathbf{M} \|_F^2 \\
&\leq \|\mathbf{W}^\star -\mathbf{L}\mathbf{L}^{\dagger}\mathbf{\mathbf{W}^\star}\mathbf{M}^{\dagger}\mathbf{M}  \|_F^2 \\
&\quad\quad  + \|\mathbf{L}\mathbf{L}^{\dagger}\mathbf{\mathbf{W}^\star}\mathbf{M}^{\dagger}\mathbf{M} - \mathbf{L} \mathbf{E} \mathbf{G}_{[N;k]}\mathbf{F}\mathbf{M} \|_F^2
\label{eq:two_term_error}
\end{aligned}
\end{equation}

Given fixed ranks, $\{R_i\}_{i=1}^N$, the first term in Eq. (\ref{eq:two_term_error}) is the irreducible error from training, which lies outside the subspace characterized by the frozen non-diagonal factor matrices. The second term in Eq. (\ref{eq:two_term_error}) is the in-subspace error which can be minimized by training $\mathbf{E}$ and $\mathbf{F}$. We can also write the following inequality for the second term in Eq. (\ref{eq:two_term_error})
\begin{equation}
\begin{aligned}
& \min_{\mathbf{E},\mathbf{F}} \|\mathbf{L}\mathbf{L}^{\dagger}\mathbf{\mathbf{W}^\star}\mathbf{M}^{\dagger}\mathbf{M} - \mathbf{L} \mathbf{E} \mathbf{G}_{[N;k]}\mathbf{F}\mathbf{M} \|_F^2\\
&= \min_{\mathbf{E},\mathbf{F}}\|\mathbf{L}(\mathbf{L}^{\dagger}\mathbf{\mathbf{W}^\star}\mathbf{M}^{\dagger} -  \mathbf{E} \mathbf{G}_{[N;k]}\mathbf{F})\mathbf{M} \|_F^2 \\
& \leq  \min_{\mathbf{E},\mathbf{F}}\|\mathbf{L}\|_F^2 \|\mathbf{M} \|_F^2 \|\mathbf{L}^{\dagger}\mathbf{\mathbf{W}^\star}\mathbf{M}^{\dagger} -  \mathbf{E} \mathbf{G}_{[N;k]}\mathbf{F})\|_F^2
\label{eq:second_term_error}
\end{aligned}
\end{equation}

Additionally, since $\mathbf{E}$ and $\mathbf{F}$ are both diagonal matrices, $\mathbf{E} \mathbf{G}_{[N;k]}\mathbf{F} =  \mathbf{G}_{[N;k]} \odot\left(\operatorname{diag}(\mathbf{E})\circ\operatorname{diag}(\mathbf{F})\right)$, where $\odot$ and $\circ$ denotes the Hadamard product and the outer-product, respectively. The term, $\mathbf{E} \mathbf{G}_{[N;k]}\mathbf{F}$, can be written as $\mathbf{G}_{[N;k]} \odot \operatorname{diag}(\mathbf{E}) \circ \operatorname{diag}(\mathbf{F})$, where $\odot$ denotes the Hadamard product, $\circ$ denotes the outer product, and $\operatorname{diag}(\cdot)$ is the vector containing the diagonal of the input matrix. Thus, we can rewrite Eq. (\ref{eq:second_term_error}) as
\begin{equation}
\begin{aligned}
&  \min_{\mathbf{E},\mathbf{F}} \|\mathbf{L}\|_F^2 \|\mathbf{M} \|_F^2 \|\mathbf{L}^{\dagger}\mathbf{\mathbf{W}^\star}\mathbf{M}^{\dagger} -  \mathbf{E} \mathbf{G}_{[N;k]}\mathbf{F})\|_F^2 \\
& \leq \min_{\mathbf{E},\mathbf{F}} g_{max}^2 \|\mathbf{L}\|_F^2 \|\mathbf{M} \|_F^2\\
& \quad \ \|\mathbf{L}^{\dagger}\mathbf{\mathbf{W}^\star}\mathbf{M}^{\dagger}\oslash\mathbf{G}_{[N;k]}-\operatorname{diag}(\mathbf{E}) \circ \operatorname{diag}(\mathbf{F})\|_F^2  
\label{eq:point_div_second_term}
\end{aligned}
\end{equation}
where $g_{max}$ is the largest entry in $\mathcal{G}$, and $\oslash$ denotes element-wise division. Denote $\mathbf{Z} = \mathbf{L}^{\dagger} \mathbf{\mathbf{W}^\star} \mathbf{M}^{\dagger} \oslash \mathbf{G}_{[N;k]}$, to arrive at
\begin{equation}
\begin{aligned}
\label{eq:rank1_approx}
& \min_{\mathbf{E},\mathbf{F}} \|\mathbf{Z}-\operatorname{diag}(\mathbf{E}) \circ \operatorname{diag}(\mathbf{F})\|_F^2  \\
& = \min_{\{\mathbf{d}^{(i)} \}_{i=1}^N} \| \operatorname{Fold}_{[N;k]}(\mathbf{Z})- \mathbf{d}^{(1)} \circ\\
&\quad \quad \quad  \mathbf{d}^{(2)}\circ \cdots \circ \mathbf{d}^{(N)}\|_F^2
\end{aligned}
\end{equation}

Let $\operatorname{Fold}_{[N;k]}(\mathbf{\hat{Z}})$ be the best rank-$1$ approximation of $\operatorname{Fold}_{[N;k]}(\mathbf{Z})$. Then, Eq. (\ref{eq:rank1_approx}) is equivalent to the minimization of the approximation error between $\operatorname{Fold}_{[N;k]}(\mathbf{\hat{Z}})$ and $\operatorname{Fold}_{[N;k]}(\mathbf{Z})$, which is given by \cite{de2000best} as
\begin{equation}
\begin{aligned}
&\min_{\{\mathbf{d}^{(i)} \}_{i=1}^N} \| \operatorname{Fold}_{[N;k]}(\mathbf{Z})-\operatorname{Fold}_{[N;k]}(\mathbf{\hat{Z}})\|_F^2\\
&= \| \operatorname{Fold}_{[N;k]}(\mathbf{Z})\|_F^2 - \| \operatorname{Fold}_{[N;k]}(\mathbf{Z})\|_2^2
\label{eq:rank_1_cp_bound}
\end{aligned}
\end{equation}
where $\|\cdot\|_2^2$ denotes the tensor spectral norm  \cite{tomioka2014spectral}. More specfically, $\|\operatorname{Fold}_{[N;k]}(\mathbf{Z})\|_2^2$ is defined as,
\begin{equation}
\begin{aligned}
&\|\operatorname{Fold}_{[N;k]}(\mathbf{Z})\|_2^2  \\
&=\sup_{\{\mathbf{u}^{(i)}\}_{i=1}^N}\operatorname{Fold}_{[N;k]}(\mathbf{Z}) \times_1 \mathbf{u}^{(1)} \times_2 \cdots \times_N \mathbf{u}^{(N)}
\end{aligned}
\end{equation}
where $\{\mathbf{u}^{(i)} \}_{i=1}^N$ are unit-norm vectors.

Combining Eq. (\ref{eq:two_term_error}),  (\ref{eq:second_term_error}), (\ref{eq:point_div_second_term}), and (\ref{eq:rank_1_cp_bound}), we arrive at
\begin{equation}
\begin{aligned}
& \min_{\{\operatorname{diag}(\mathbf{d}^{(i)}) \}_{i=1}^N} \| \mathbf{\mathbf{W}^\star} - \mathbf{W}_{TeRA}\|_F^2 \\
& \leq \|\mathbf{\mathbf{W}^\star}- \mathbf{L}\mathbf{L}^{\dagger}\mathbf{\mathbf{W}^\star}\mathbf{M}^{\dagger}\mathbf{M}\|_F^2 \\
& \ + g_{max}^2(\|\mathbf{Z}\|_F^2 - \|\operatorname{Fold}_{[N;k]}(\mathbf{Z})\|_2^2) \|\mathbf{L}\|_F^2\|\mathbf{M}\|_F^2.\
\end{aligned}
\end{equation}

\end{proof}

\section{Ablation Study on TeRA Placements in Transformers}
We explored different placements of TeRA on the attention weight matrices. Specifically, we compared three cases: 1) Applying TeRA only on the query matrices; 2) only on the value matrices; 3) on both the query and the value matrices. As shown in Table \ref{tab:q,v,qv}, applying TeRA on both the query and the value matrices yields the best performance.

\begin{table*}[ht]
\centering
\footnotesize 
\begin{adjustbox}{max width=\textwidth}
\begin{tabular}{llc|cccccccccc}
\toprule
\textbf{Model} & \textbf{Method} & \textbf{Params (\%)} & \textbf{BoolQ} & \textbf{PIQA} & \textbf{SIQA} & \textbf{ARC-c} & \textbf{ARC-e} & \textbf{OBQA} & \textbf{HellaS} & \textbf{WinoG} & \textbf{Average} \\
\midrule
\multirow{6}{*}{Llama-2-7B}
 & LoRA ($r$=32) & 0.2484 & 67.65 & 79.22 & 78.20 & 69.20 & 83.88 & 78.60 & 81.05 & 80.98 & 77.35 \\
 & LoRA ($r$=128) & 0.9861 & 69.76 & 83.62 & 79.38 & 72.61 & 86.53 & 81.80 & 88.57 & 84.77 & 80.88 \\
 & HiRA ($r$=1) & 0.0078 & 65.35 & 77.97 & 72.42 & 62.03 & 81.48 & 64.80 & 79.90 & 70.01 & 71.74 \\
 & HiRA ($r$=32) & 0.2484 & 69.39 & 83.24 & 78.86 & 71.33 & 86.57 & 81.40 & 87.23 & 81.69 & 79.97 \\
 & HiRA ($r$=128) & 0.9861 & 68.59 & 82.43 & 80.04 & 71.42 & 85.94 & 79.80 & 87.30 & 81.37 & 79.61 \\
 & \textbf{TeRA (Ours)} & \textbf{0.0039} & 65.78 & 81.23 & 78.92 & 69.45 & 84.05 & 81.00 & 85.94 & 82.64 & 78.63 \\
\midrule
\multirow{6}{*}{Llama-3-8B} 
 & LoRA ($r$=32) & 0.1695 & 71.99 & 85.91 & 79.58 & 76.19 & 88.55 & 82.60 & 92.54 & 85.63 & 82.88 \\
  & LoRA ($r$=128) & 0.6744 & 73.61 & 87.87 & 82.55 & 81.74 & 92.21 & 87.8 & 95.59 & 88.24 & 86.20 \\
 & HiRA ($r$=1) & 0.0053 & 68.04 & 85.75 & 75.54 & 76.96 & 90.32 & 77.00 & 89.04 & 77.43 & 80.01 \\
 & HiRA ($r$=32) & 0.1695 & 73.09 & 88.85 & 81.06 & 80.38 &  92.68 & 86.20 & 94.37 & 85.87 & 85.31 \\
 & HiRA ($r$=128) & 0.6744 & 74.31 & 90.10 & 81.83 & 82.51 & 92.97 & 88.4 & 95.92 & 88.16 & 86.78 \\
& \textbf{TeRA (Ours)} & \textbf{0.0033} & 70.70 & 88.08 & 81.58 & 80.89 & 92.00 & 88.00 & 94.92 & 86.27 & 85.31 \\
\bottomrule
\end{tabular}
\end{adjustbox}
\caption{Performance and parameter efficiency of TeRA compared with higher-rank LoRA/HiRA.}
\label{tab:common_sense_higher_rank}
\end{table*}

\begin{table*}[ht]
\centering
\footnotesize 
\begin{adjustbox}{max width=\textwidth}
\begin{tabular}{ll|cccccccccc}
\toprule
\textbf{Model} & \textbf{Method}  & \textbf{BoolQ} & \textbf{PIQA} & \textbf{SIQA} & \textbf{ARC-c} & \textbf{ARC-e} & \textbf{OBQA} & \textbf{HellaS} & \textbf{WinoG} & \textbf{Average} \\
\midrule
\multirow{2}{*}{Llama-2-7B}
 & TeRA  & \textbf{65.78} & \textbf{81.23} & \textbf{78.92} & \textbf{69.45} & \textbf{84.05} & \textbf{81.00} & \textbf{85.94} & \textbf{82.64} & \textbf{78.63} \\
 & TeRA$_{unique}$  & 61.47 & 0.16 & 0 & 0.09 & 0.04 & 0 & 0.15 & 0 & 7.74 \\
\midrule
\multirow{2}{*}{Llama-3-8B} 
& TeRA & \textbf{70.70} & \textbf{88.08} & \textbf{81.58} & \textbf{80.89} & \textbf{92.00} & \textbf{88.00} & \textbf{94.92} & \textbf{86.27} & \textbf{85.31} \\
& TeRA$_{unique}$  & 58.47 & 66.27 & 29.99 & 16.30 & 14.35 & 3.60 & 11.86 & 11.92 & 26.59 \\
\bottomrule
\end{tabular}
\end{adjustbox}
\caption{Performance comparison of TeRA and TeRA$_{unqiue}$, where the frozen Tucker core and factor matrices are unique for each finetuned matrix in each layer.}
\label{tab:common_sense_tera_unique}
\end{table*}

\section{Hyperparameters}
All experiments were conducted using one NVIDIA A100 (80GB) GPU. The AdamW optimizer \cite{loshchilov2018decoupled} was employed with 100 warm-up steps. The general hyperparameters for TeRA for all three datasets mentioned in the main paper are listed in Table \ref{tab:hyperparam}. The frozen factors in TeRA are initialized with Kaiming init \cite{he2015delving}.

\begin{table}[ht]
\centering
\begin{tabular}{ll}
\toprule
\textbf{Hyperparameter} & \textbf{Value} \\
\hline
Optimizer & AdamW \\
Weight Decay & 0 \\
Base Model & [Llama-2-7B, Llama-3-8B] \\
Learning Rate & [1e-3, 2e-4, 3e-4] \\
Warm Up & 100 steps \\
Batch Size & 32 \\
Target Modules & q\_proj, v\_proj \\
Evaluation Steps & Every 80 steps \\
indim & [2,4,8,16,64,4096] \\
outdim & [2,4,8,16,64,4096] \\
\bottomrule
\end{tabular}
\caption{Hyperparameters for TeRA.}
\label{tab:hyperparam}
\end{table}

The dimension used for splitting the two models across the three datasets were chosen from the last two rows in Table \ref{tab:hyperparam}, where \texttt{indim} represents the dimension of which the input feature is split into, while \texttt{outdim} denotes the dimension of which the output feature is split into. For example, in \texttt{indim=4096} and \texttt{outdim=8} setting, the weight update matrix of shape (4096,4096) is unfolded from a tensor of shape (4096,8,8,8,8).

For the other methods reported in the paper, we chose $r$ according to the recommendations in the original paper. Specifically, LoRA uses a rank of 32; LoRETTA employs LoRETTA$_\text{rep}$ with a base rank of 32 and a TT-rank of 8; and VeRA uses a rank of 256.

\section{Computational Resource Details} 
All experiments were conducted using one NVIDIA A100 (80GB) GPU on a Linux operation system. Linux operating system was used. Pytorch version used was 2.2.1. We used Tranformers 4.41.1, and PEFT 0.17.1 from Huggingface.

\section{Statistical Significances}
Wilcoxon signed-rank test \cite{c4091bd3-d888-3152-8886-c284bf66a93a} was used to test whether TeRA statistically significantly outperforms methods with similar number of parameters, such as HiRA ($r=1$) \cite{huang2025hira}, VeRA \cite{kopiczko2024vera} and LoRETTA \cite{yang2024loretta}, across the three datasets tested. TeRA performed statistically better than HiRA ($r=1$) with a p-value of $3.97\times 10^{-7}$. TeRA performed statistically better than VeRA with a p-value of $3.97\times 10^{-7}$.
TeRA performed statistically better than LoRETTA with a p-value of $2.20 \times 10^{-5}$. TeRA also performed statistically better than the LoRA with a p-value of $2.37 \times 10^{-4}$. We also used Wilcoxon signed-rank test to test whether HiRA ($r=32$) performed statistically better than TeRA. The results confirm that HiRA ($r=32$) did not perform statistically better or worse than TeRA.

\begin{table*}[ht]
\centering
\begin{adjustbox}{max width=\textwidth}
\begin{tabular}{lll|cccccccccc}
\toprule
\textbf{Model} & \textbf{Method} & \textbf{Params (\%)} & \textbf{BoolQ} & \textbf{PIQA} & \textbf{SIQA} & \textbf{ARC-c} & \textbf{ARC-e} & \textbf{OBQA} & \textbf{HellaS} & \textbf{WinoG} & \textbf{Average} \\
\midrule
\multirow{4}{*}{Llama-3-8B} 
& TopLoRA ($r=16$) & 0.1369 & \textbf{72.60} & \textbf{88.30} & 80.09 & \textbf{81.14} & \underline{91.92} & \underline{85.20} & \underline{94.31} &	\textbf{86.42} & \underline{85.00} \\
& PiSSA ($r=16$) & \underline{0.0848} &	67.33 &	85.09 &	79.84 &	75.42 &	88.55 &	80.60 &	91.96 &	85.24 &	81.76 \\
& DoRA ($r=32$) & 0.1715 & 58.65	& 85.36 & \underline{80.45} & 76.11 & 89.27 & 82.8 & 92.33 &	85.56 &	81.32 \\
& TeRA & \textbf{0.0033} & \underline{70.70} & \underline{88.08} & \textbf{81.58} & \underline{80.89} & \textbf{92.00} & \textbf{88.00} & \textbf{94.92} & \underline{86.27} & \textbf{85.31} \\
\bottomrule
\end{tabular}
\end{adjustbox}
\caption{Performance comparison of TopLoRA, PiSSA, DoRA, and TeRA.}
\label{tab:common_sense_tera_more_comparison}
\end{table*}

\section{LoRA and HiRA with higher ranks}

To highlight the superior parameter efficiency and accuracy trade-off of TeRA, we evaluated LoRA and HiRA with higher ranks. The results in Table \ref{tab:common_sense_higher_rank} show that TeRA achieved comparable performance, while requiring about $200\times$ fewer trainable parameters. 

\section{Ablation on Sharing Tucker Core and Factor Matrices}
\label{sec:non-sharing}
To illustrate the necessity of using a shared Tucker core and factor matrices across all layers, we compared TeRA with its variant TeRA$_{unique}$, where a unique frozen Tucker core $\mathcal{G}$ and factor matrices $\{\mathbf{A}^{(i)}\}_{i=1}^N$ were used for each fine-tuned matrix in each layer. As shown in Table \ref{tab:common_sense_tera_unique}, a significant performance degradation is observed for TeRA$_{unique}$ under the same experimental setup as TeRA. 

This degradation might be due to the loss of a consistent subspace across layers. When the Tucker core and factor matrices are shared, each layer projects its weight update into the same subspace through its trainable diagonal matrices. As a result, the updates from different layers can reinforce or complement each other. In contrast, without sharing, the updates in each layer are confined to independent random subspaces. Since only the diagonal matrices are trainable, each layer must search for effective directions in isolation, which reduces efficiency and makes optimization more difficult.

\section{Rank Comparison between High-Rank Adapters}
\label{sec:rank_com}
\begin{figure}[ht]
    \centering
    \includegraphics[width=0.48\textwidth]{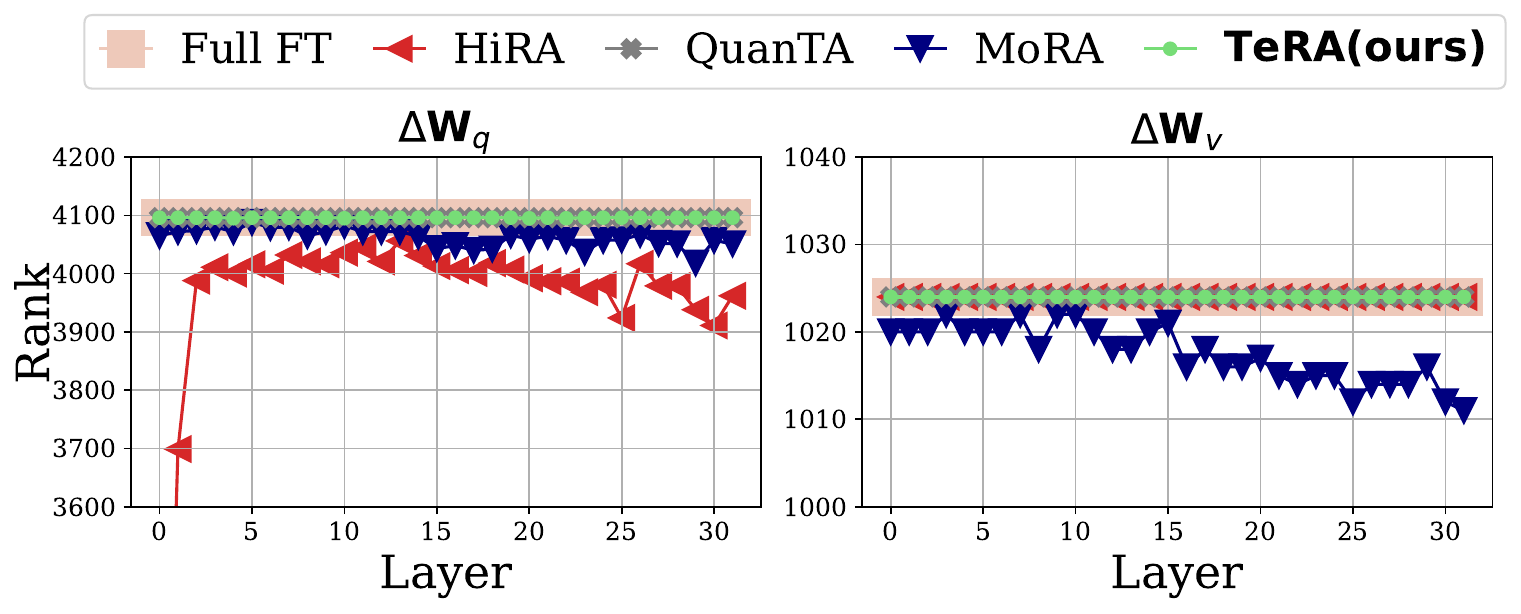}
    \caption{Rank analysis of $\Delta \mathbf{W}_q$ (max allowed rank of $4096$) and $\Delta \mathbf{W}_v$ (max allowed rank of $1024$) across Llama-3-8B layers for high-rank adapters.}
    \label{fig:rank_highrank_methods_appen}
\end{figure}

Figure \ref{fig:rank_highrank_methods_appen} shows the detailed comparison of ranks among high-rank methods.

\section{Additional experiments}

The performance comparison of more PEFT methods, TopLoRA \cite{li2025beyond}, such as PiSSA \cite{meng2024pissa} and DoRA \cite{liu2024dora}, is illustrated in Table \ref{tab:common_sense_tera_more_comparison}. TeRA achieves the best overall performance, while using the least number of trainable parameters, demonstrating again the superior performance and parameter-efficeincy trade-off.

\section{Artifact License and Others}
Commonsense170k uses Apache-2.0 License. ConvAI2 uses MIT license. Math10k uses Apache-2.0 License. We used AI assistants to help polish up the writing.

\end{document}